\documentclass[conference]{IEEEtran}

\usepackage{cite}
\usepackage{amsmath,amssymb,amsfonts}
\usepackage{algorithm}
\usepackage{algpseudocode}
\usepackage{graphicx}
\usepackage{booktabs}
\usepackage{hyperref}
\usepackage{textcomp}
\usepackage{xcolor}
\usepackage{tikz}
\usepackage{pgfplots}
\usepackage{commands}
\usepackage{multirow}
\usepackage[bottom]{footmisc}

\usepackage{bbm}

\usepackage{pdfpages}

\usepackage{array}
\newcolumntype{V}{>{\hspace{-2\tabcolsep}\savebox{\bitbucket}\bgroup}{c}<{\egroup}}
\definecolor{dgreen}{rgb}{0.0,0.6,0.0}
\definecolor{dred}{rgb}{0.6,0.0,0.0}

\newcommand{\fn}[1]{~\scriptsize\textcolor{dgreen}{#1}}
\newcommand{\fd}[1]{~\scriptsize\textcolor{dred}{#1}}

\hypersetup{
  colorlinks,
  citecolor=blue,
  linkcolor=black,
  urlcolor=black}

\def\BibTeX{{\rm B\kern-.05em{\sc i\kern-.025em b}\kern-.08em
    T\kern-.1667em\lower.7ex\hbox{E}\kern-.125emX}}

\pgfplotsset{compat=1.17}
\begin{document}

\title{Alternating Objectives Generates Stronger PGD-Based Adversarial Attacks}

\author{\IEEEauthorblockN{Antoniou Nikolaos \IEEEauthorrefmark{1}, Efthymios Georgiou \IEEEauthorrefmark{1}\IEEEauthorrefmark{2}, Alexandros Potamianos \IEEEauthorrefmark{1}}

\IEEEauthorblockA{\IEEEauthorrefmark{1} School of Electrical and Computer Engineering, National Technical University of Athens, Athens, Greece\\
\IEEEauthorrefmark{2} Institute for Language and Speech Processing, Athena Research Center, Athens, Greece\\ 
antoniou\_nikos@hotmail.com, efthygeo@mail.ntua.gr, potam@central.ntua.gr}
 }
\maketitle

\begin{abstract}
Designing powerful adversarial attacks is of paramount importance for the evaluation of $\ell_p$-bounded adversarial defenses.  Projected Gradient Descent (PGD) is one of the most effective and conceptually simple algorithms to generate such adversaries. The search space of PGD is dictated by the steepest ascent directions of an objective. Despite the plethora of objective function choices, there is no universally superior option and robustness overestimation may arise from ill-suited objective selection. Driven by this observation, we postulate that the combination of different objectives through a simple loss alternating scheme renders PGD more robust towards design choices. We experimentally verify this assertion on a synthetic-data example and by evaluating our proposed method across 25 different $\ell_{\infty}$-robust models and 3 datasets. The performance improvement is consistent, when compared to the single loss counterparts. In the CIFAR-10 dataset, our strongest adversarial attack outperforms all of the white-box components of AutoAttack (AA) ensemble \cite{croce2020reliable}, as well as the most powerful attacks existing on the literature, achieving state-of-the-art results in the computational budget of our study ($T=100$, no restarts).
\end{abstract}


\section{Introduction}\label{introduction}

The advent of Deep Learning (DL) caused a paradigm shift and revolutionized the way that various interesting applications are approached. Such a wide adoption, however, demands from the research community to comprehend the scenarios where Deep Neural Networks (DNNs) malfunction. This necessity becomes even more imperative when considering the abundance of safety-critical applications that do not leave room for complacency, e.g., autonomous driving. Unfortunately, DNNs have significant failure modes and behave counterintuitively. A prominent instance of this behaviour is illustrated by Szegedy et al. \cite{szegedy2014intriguing}, where they showcase that DNN-based image classifiers are vulnerable against \emph{adversarial examples}. These examples arise from applying humanly imperceptible perturbations to clean images, which are capable of degrading the model's predictive performance.  This finding triggered research interest on two fronts: \emph{Adversarial Attacks}, which are algorithms to generate such malicious examples and \emph{Adversarial Defenses}, which are methods of increasing the robustness of neural networks. Adversarial robustness is primarily studied through the $\lp$-bounded threat model, where the perturbation's $\lp$-norm is bounded by a small constant. 

The robustness of Adversarial Defenses, on a given dataset, is estimated by the rate of test set's adversarial examples that the defense can properly classify. Of course, the estimated rate (also called robust accuracy) depends on the strength of the attacking algorithm that will be used for evaluation. Employing weak attacks to evaluate robustness creates a false sense of security, an issue widely known as robustness overestimation \cite{athalye2018obfuscated,uesato2018adversarial,tramer2020adaptive}. 

Arguably, Projected Gradient Descent (PGD) is the most popular adversarial attack used to evaluate $\lp-$bounded robustness. PGD operates by iteratively following the steepest ascent directions of an objective function, often called the surrogate. PGD has raised in many guises in the adversarial attack literature:  Goodfellow et al. \cite{goodfellow2015explaining} propose to attack networks through the Fast Gradient Sign Method (FGSM), which takes a single normalized step, i.e., applying the sign function in the case of $\linf$-norm, towards the steepest ascent direction. Kurakin et al. \cite{kurakin2017adversarial} demonstrate that the multi-step variants of FGSM are capable of producing significantly stronger attacks. Dong et al. \cite{dong2018boosting} suggest a modification of the iterative FGSM that integrates a momentum term. Madry et al. \cite{madry2019deep} link the iterative FGSM with the classical optimization algorithm of PGD.

Despite that PGD combines both simplicity (in terms of implementation) and strength, it has been shown that its performance can be hindered by ill-suited selection of hyperparameters, e.g., fixed step size \cite{croce2020reliable}. Another hyperparameter of consideration is the surrogate loss, for which literature has converged into 3 options: Cross-Entropy (CE) \cite{goodfellow2015explaining,madry2019deep}, Margin (a.k.a. CW) loss \cite{carlini2017towards} and the Difference of Logits Ratio (DLR) loss \cite{croce2020reliable}, with the appealing property of scale-invariance. However, empirical evidence (e.g., as in Figures 9-11 of \cite{croce2020reliable}) shows that there is no universally superior objective and its effectiveness depends on the architecture, weights, training dataset etc. On top of this, certain choices may be improper in special problematic cases: 1) CE yields zero gradients for inputs where the classifier assigns the entire probability mass to the ground truth class \cite{carlini2017towards,croce2020reliable}, 2) both CE and CW are not scale-invariant hence logit rescalings may induce gradient masking \cite{croce2020reliable} and 3) Ma et al. \cite{Ma2020ImbalancedGA} assert that objectives which involve multiple logit terms, i.e., all three of CE,CW and DLR, may suffer from the problem of \emph{gradient imbalance} where logits have quite disparate magnitudes and one term alone steers the optimization trajectory towards non-optimal solutions. 

In this work, PGD is studied from the perspective of surrogate loss. In order to alleviate potentially weak PGD performance arising from poor surrogate selection, we propose to combine different objectives in the same run of PGD. Hopefully, this combination will render PGD less dependent to the surrogate hyperparameter. We identify that a simple alternation of objectives during PGD is sufficient to induce significant boost on the PGD performance over the single loss variants. Further qualitative analysis implies that the switching between different objectives helps the algorithm to expand its search space, visiting more distant intermediate points during its execution. 

\noindent
In this paper, we make the following key contributions:  
\begin{itemize}
    \item We propose to combine multiple objectives during PGD through alternating between them during optimization, in order to alleviate potential flaws of each objective. Our proposed strategy outperforms the single-loss PGD variants in 25 out of 25 (15 on CIFAR-10, 6 on CIFAR-100 and 4 on ImageNet) tested $\linf-$bounded robust models. 
    \item For the CIFAR-10 dataset, our attack outperforms the three white-box components of AutoAttack \cite{croce2020reliable}: $\apgdce$, $\apgddlr$ and FAB attack \cite{Croce2020MinimallyDA}. Furthermore, in most cases our attack achieves higher Attack Success Rate (ASR) than the strongest baselines (for $T=100$ iterations and $R=1$ restart) in the literature: GAMA-PGD \cite{Sriramanan2020GuidedAA} and MD attack \cite{Ma2020ImbalancedGA}. 
    \item We present extensive experimentation and analysis regarding the proposed alternation scheme, including: 1) A synthetic example which highlights how PGD with a single loss can fail, 2) Qualitative analysis indicating that switching losses promotes search diversity and 3) Ablation experiments which demonstrate that this loss combination strategy is more effective than two other combining methods. 
\end{itemize}

\noindent
The remainder of this paper is organized as follows: Section \ref{background} provides the necessary background, covering basic aspects of the worst-case $\lp-$bounded adversarial robustness, Section \ref{sec:relatedwork} briefly discusses research work related to PGD-based attacks, since PGD is the main topic of our study. In Section \ref{sec:experiments} we conduct numerous experiments to verify the effectiveness of our proposed method, whereas in Section \ref{sec:conclusion} we discuss how our study differs from previous related work.

\section{Background}\label{background}

\subsection{Notation}

Image-label pairs are denoted as $(\vx,y) \in \cX \times \cY$ where $ \cX \subseteq \bR^D, \cY \subseteq \mathbb{Z}$. The classifier's logit representation will be denoted as $\vz(\vx) \in \mathbb{R}^C$ (or simply $\vz$), where $C$: the total number of classes. Applying a softmax layer to the logit vector produces the probability vector $p(y|\vx)$. The classification decision will be denoted as $f(\vx)$, hence $f(\vx) = \argmax{i \in [C]}\ \vz(\vx)_i$, where $[C] = \{1,...,C\}$. The surrogate loss $\cL(\vz(\vx),y)$ (which will also be referred as $\cL(\vx,y)$, for brevity's sake), e.g., cross-entropy, measures the model's ability to assign the label $y$ to example $\vx$. 

\subsection{Threat Model}
The constraint of creating an imperceptible perturbation is approximated through the bounded $\lp-$norm condition. The generation of adversarial attacks should obey this restriction, returning an output that lies within the $\lp-$ball of radius $\epsilon$ around the clean input $\vx$. Hence, the search space of potential adversaries for the image $\vx$ can be expressed as:
\begin{equation}
    \mathcal{S}(\vx) = \{ \vx': \|\vx-\vx'\|_p \leq \epsilon \}
\end{equation}
Despite that the $\lp-$bounded threat model is only a crude approximation of true similarity between data samples like images, solving the problem of $\lp-$bounded robustness can be viewed as an important stepping stone towards confronting more realistic scenarios.

\subsection{A taxonomy of $\lp-$bounded adversarial attacks}

Next we present a basic categorization of adversarial attacks based on their capabilities during generation and their end goal.\\

\noindent
\textbf{Adversary's Knowledge.} Based on the amount of information that the adversary has at its disposal, attacks can be divided into two major categories: \emph{white-box} and \emph{black-box}. In the former, the attacker has access to every aspect of the model: its architecture, weights and training data. This allows the adversary to obtain the network's gradients w.r.t. the input which is particularly useful when creating attacks. In the latter category, however, the adversary can only use the model as an oracle, feeding an input point and getting access to the output vector, or sometimes just to the output class. 

Despite that typical real-world scenarios are more similar to the black-box setting, white-box attacks constitute a much more stronger threat model. Therefore, the evaluation of adversarial defenses is typically performed based on white-box attacks. \\

\noindent
\textbf{Low Confidence vs Low Distortion.} Attacks are also divided into minimum-confidence and minimum-norm. In the former, the attack algorithm is based on the following formulation, for the input-label pair $(\vx,y)$:
\begin{equation}
    \vect{\delta}: \argmax{\vect{\delta}}~ \cL_{0/1}(f(\vx+\vect{\delta}),y) \ \text{s.t.} \ \vx+\vect{\delta} \in \mathcal{S}(\vx)
\end{equation}
where $\cL_{0/1}(f(\vx),y) = \ind[f(\vx)\neq y]$ is the 0-1 loss, which due to its discontinuity is replaced by some surrogate loss $\cL$ such as cross-entropy. These attacks aim to reduce the ground truth label's confidence as much as possible by spending the entire attack budget $\epsilon$, hence they typically lie on the boundary surface of the feasible set $\mathcal{S}$. The most prominent examples of minimum-confidence adversarial attacks is the Fast Gradient Sign Method (FGSM) \cite{goodfellow2015explaining}, the Iterative-FGSM \cite{kurakin2017adversarial} and Projected Gradient Descent (PGD) \cite{madry2019deep}. 

Minimum-norm attacks aspire to find the smallest possible perturbation that leads to misclassification:
\begin{equation}
    \vect{\delta}: \argmin{\vect{\delta}} \| \vect{\delta} \|_p \ \text{s.t.} \ f(\vx+\vect{\delta}) \neq y 
\end{equation}
where $y$: the ground-truth label of $\vx$. Such attacks usually find adversaries that are within smaller $\lp$-distance from the clean input $\vx$ than the perturbation bound $\epsilon$. Popular examples of this category are: Carlini-Wagner (CW) attack \cite{carlini2017towards}, DDN attack \cite{Rony2019DecouplingDA}, Fast Minimum Norm (FMN) \cite{pintor2021fast} and Fast Adaptive Boundary (FAB) attack \cite{Croce2020MinimallyDA} among others.\\ 

\noindent 
\textbf{Untargeted vs Targeted. } Another criterion of dividing adversarial attacks is whether the adversary desires to force a specific label to the attack. In \emph{targeted} attacks, the attack is considered successful if the corresponding adversarial example is classified into a certain target class. In \emph{untargeted} attacks, the goal is simply to produce an example which is incorrectly classified, with no constraint on its new label. Usually, the transition between the two categories is as simple as slightly modifying the objective function, i.e., from descending the target label's confidence to ascending the ground-truth label's confidence.

\subsection{Empirical Adversarial Defenses}
Training $\lp-$robust neural networks, i.e., networks that are resilient against $\lp$-bounded adversarial attacks, is a complicated problem since we aspire to simultaneously realize two goals. First, the classifier is asked to perform well on unseen examples drawn from the same distribution as the examples used during training. An additional requirement is to find networks that produce smooth predictions, assigning the same label to all data residing inside the $\lp-$ball of such examples. 
The most standard way of increasing $\lp-$bounded robustness is Adversarial Training (AT) \cite{goodfellow2015explaining,madry2019deep}; in AT, the defender aims to minimize the \emph{robust expected risk}:

\begin{equation}\label{eq:robrisk}
    \mathcal{R}^{f}_{\text{rob}}(\vth) = \ExpDnew \Big [ \max_{\vect{\delta}:\|\vect{\delta}\|_p \leq \epsilon } \ind [f_{\vth} (\vx+\vect{\delta}) \neq y]\Big]
\end{equation}

The inner expression coincides with the task of finding the worst-case $\lp-$bounded adversarial example. Madry et al. \cite{madry2019deep} confront the problem through the first-order method of PGD. An important barrier of this method is the additional computational overhead. The iterative PGD process renders this method costly in terms of compute, hence a line of research aims to increase robustness using one-step adversaries \cite{Shafahi2019AdversarialTF,Wong2020FastIB,Rice2020OverfittingIA,andriushchenko2020understanding}, in order to restrain the overall training time to similar levels as with standard training. 
Another important work on adversarial defenses is the TRADES framework, introduced by Zhang et al. \cite{zhang2019theoretically}. The robust expected risk of \autoref{eq:robrisk} can be decomposed as the sum of two individual terms. The first term represents the \emph{classification error}, where the optimization searches parameters that generalize well. The other term, dubbed as \emph{boundary error}, can be considered as exerting a regularizing effect, where it imposes decision ``smoothness" between inputs inside the same $\lp-$ball. 

Schmidt et al. \cite{Schmidt2018AdversariallyRG} provide evidence that adversarially training classifiers may require an increasing amount of data. Following this, many works \cite{carmon2019unlabeled,Zhai2019AdversariallyRG,Uesato2019AreLR} explore the use of both pseudo-labeled additional data and elaborate data augmentation techniques. \\

\noindent
\textbf{Robustness Overestimation.} Evaluating the true degree of $\lp$-bounded robustness of empirical methods is intractable, since one needs to calculate the average 0-1 risk on a held-out test set. Typically, the defender deploys a strong attacking algorithm to obtain a lower bound on the true risk. However, this trial-and-error technique can provide misleading results. Failing to select a proper attacking algorithm creates an inaccurate sense of security \cite{athalye2018obfuscated,uesato2018adversarial,tramer2019adversarial}. Importantly, these works propose numerous indicators that demonstrate whether the evaluation suffers from this issue and guidelines of how to properly evaluate a defense.\\
The introduction of RobustBench \cite{robustness}, based on the AutoAttack ensemble (comprised of three white-box \cite{croce2020reliable},\cite{Croce2020MinimallyDA} and one black-box \cite{andriushchenko2020square} methods), contributed to a consensus regarding the evaluation of $\lp-$bounded robustness: A newly proposed defense is first ``passed" through an AutoAttack evaluation, and then the defender can also perform adaptive attacks \cite{tramer2019adversarial}, based on potential model-specific weaknesses. \\
Despite the general adoption of AutoAttack as the standard way to perform first-order robustness evaluations, the community is constantly exploring faster and more powerful attack ensembles \cite{liu2022practical}, \cite{yu2021lafeat}. 
\section{Related Work}\label{sec:relatedwork}

Projected Gradient Descent (PGD) \cite{madry2019deep,kurakin2017adversarial} is the most popular minimum-confidence attack. PGD has been the de facto standard for producing $\lp-$bounded adversarial attacks, especially in the case of $p=\infty$. In short, PGD can be expressed as: 
\begin{equation}
    \vx^{(t+1)} = \mathcal{P}_{\mathcal{S}(\vx)} \Big[ \vx^{(t)} + \eta^{(t)} \vect{\delta}^{(t)}\Big]    
\end{equation}
where $\vx^{(t)}$: the iterate, $\eta^{(t)}$: step size, $\vect{\delta}^{(t)}$: update rule of t-th iteration and $\mathcal{P}_{\mathcal{S}}$: the projection operation, which maps the updated iterate into the feasible region $\mathcal{S}$, which in our case is the $\lp-$ball of radius $\epsilon$ around $\vx$. Typically, this procedure is repeated multiple times from different random initializations. For a more comprehensive view of how PGD is used to generate adversarial attacks, we refer to the work of Gowal et al. \cite{Gowal2019AnAS}, where they present a ``holistic" pseudoalgorithm.

In the following discussion we present how one can manipulate the basic building blocks of PGD, namely the optimizer, step size, initialization strategy and surrogate loss, in order to improve its adversarial generation stregnth. \\

\noindent
\textbf{Optimizer.} 
The optimizer determines the form of the update rule $\vect{\delta}^{(t)}$. In its simplest version, assuming the surrogate loss $\cL(\vx,y)$, PGD follows the steepest direction of unit $\lp$-norm, e.g., the sign of $\nabla_{\vx^{(t)}} \cL(\vx^{(t)},y)$ in the case of $p=\infty$, or a simple norm-rescaling when $p=2$. In the C\&W attack \cite{carlini2017towards}, the proposed objective is optimized through Adam \cite{kingma2014adam}. The Adam optimizer has also been leveraged in PGD-based works \cite{Gowal2019AnAS,uesato2018adversarial}. Dong et al. \cite{dong2018boosting} suggested the incorporation of momentum \cite{POLYAK19641} in the PGD update rule. Subsequently, Croce and Hein \cite{croce2020reliable} proposed the AutoPGD (APGD) variant, wherein the update term is augmented by momentum. Yamamura et al. \cite{yamamura2022diversified} developed the Auto Conjugate Gradient (ACG) method, which is an elaborate optimizer, adjusting the update rule based on accumulated gradient information from previous steps. ACG is experimentally shown to outperform APGD for a sizable collection of robust models. \\

\noindent
\textbf{Step Size.} Another hyperparameter which affects the performance of PGD is the step size $\eta^{(t)}$. In early works, its value is held constant during the entire optimization procedure, e.g., to $\alpha=\epsilon/4$ for $\linf$-attacks in CIFAR-10. Croce and Hein \cite{croce2020reliable} conduct large-scale experiments regarding the optimal fixed value, but one immediate corollary is that it greatly depends on the model. Generally, the common trend is to perform some kind of scheduling, where the step size is gradually reduced over time: In \cite{Gowal2019AnAS}, \cite{Sriramanan2020GuidedAA}, the authors apply ten-fold drops at two intermediate timesteps; Ma et al. \cite{Ma2020ImbalancedGA} propose a cosine-annealing scheme, where the step size decays from $2\epsilon$ to 0. In their recent work, Liu et al. \cite{liu2022practical} adopt a similar decaying strategy. Another interesting way of manipulating this hyperparameter is as in the AutoPGD method \cite{croce2020reliable}; They initially set it to a large value $\alpha=2\epsilon$, in order to explore the search space sufficiently well. Then, as the optimization proceeds and the iterate gets closer to some local optimum, the need of a more localized search calls for smaller step sizes. Hence, it is halved in specific checkpoints, according to the optimization progress, i.e., based on whether the objective function is reducing or not. 
\\

\noindent
\textbf{Initialization.} Proper initialization plays also a crucial role in the final performance. Typically, the initial point $\vx^{(0)}$ can be either set to the clean image $\vx$, or alternatively, random noise may be added to the clean image: $\vx^{(0)} = \vx + \vect{\zeta}$, where $\vect{\zeta}$ is drawn from some noise distribution. The attack is then repeated multiple times, initialized from different starting points. Tashiro et al. \cite{Tashiro2020OutputDI} suggest that random initialization may lead to starting points with nearly identical output space representations, hence the attack generates similar results even if executed for many restarts.  Output Diversified Initialization (ODI) \cite{Tashiro2020OutputDI} counteracts this by maximizing the similarity of starting point's logit vector with a random output direction, in the first few PGD iterations. Recently, Liu et al. \cite{liu2022practical} introduced Adaptive AutoAttack ($\text{A}^3$), the new state-of-the-art attack ensemble. $\text{A}^3$ uses an  adaptive initialization strategy, where the starting points are generated by ODI, but instead of following random output space direction, the vector is selected according to prior knowledge of perturbations that led to misclassification. \\

\noindent
\textbf{Surrogate Loss.} The maximization of 0-1 loss is intractable for complex function classes as those represented by deep neural networks \cite{ARORA1997317}. It is common to substitute it with a surrogate, differentiable loss which is amenable to optimization methods. A natural candidate is the cross-entropy (CE) objective, which coincides with the negative log-likelihood of the ground truth class. In their seminal work, Carlini and Wagner \cite{carlini2017towards} tested various formulations, obtaining the best performance for the so called margin (or CW) loss. A shared defect in both of these objectives is the lack of scale-invariance, which may be translated in deteriorated performance due to gradient masking. Croce and Hein \cite{croce2020reliable} introduce the Difference of Logits Ration (DLR) loss, which rescales the margin loss to acquire the property of scale-invariance. Most of the literature involves these three options, whose expressions are included below for completeness:

\begin{equation}\label{eq:losses}
    \begin{split}
        \text{CE}(\vx,y) &= - \log p(y|\vx) = - \vz_y + \log \sum_{j=1}^C \exp(\vz_j) \\
        \text{CW}(\vx,y) &= - \vz_y + \max_{j \neq y} \vz_j \\
        \text{DLR}(\vx,y) &= - \frac{\vz_y + \max_{j \neq y} \vz_j }{\vz_{\pi_1}-\vz_{\pi_3}} 
    \end{split}
\end{equation}
where $\vz_{\pi}$: the logit vector sorted in descending order. Gowal et al. \cite{Gowal2019AnAS} propose the MultiTargeted PGD variant which divides the iteration budget into runs of equal size, where each run optimizes the targeted margin loss, for a different target label per run. Their experiments indicate that the MultiTargeted strategy exploits more judiciously the given computational budget.
Sriramanan et al. \cite{Sriramanan2020GuidedAA} augment the standard margin loss expression with a regularization term which is set to the MSE between the logit vector of the adversary and its clean counterpart. The weighting coefficient of MSE term is gradually decayed to zero.  Ma et al. \cite{Ma2020ImbalancedGA}, in an effort to address the issue of imbalanced gradients, optimize only one of the two margin loss terms for the first half of iterations before switching to the typical expression which contains both terms. In the next restart, they repeat the process by using the other term for the first stage of optimization. 
\section{Methodology}\label{section:method}

\begin{figure*}[!ht]
    \centering
    \includegraphics[width=0.45\textwidth]{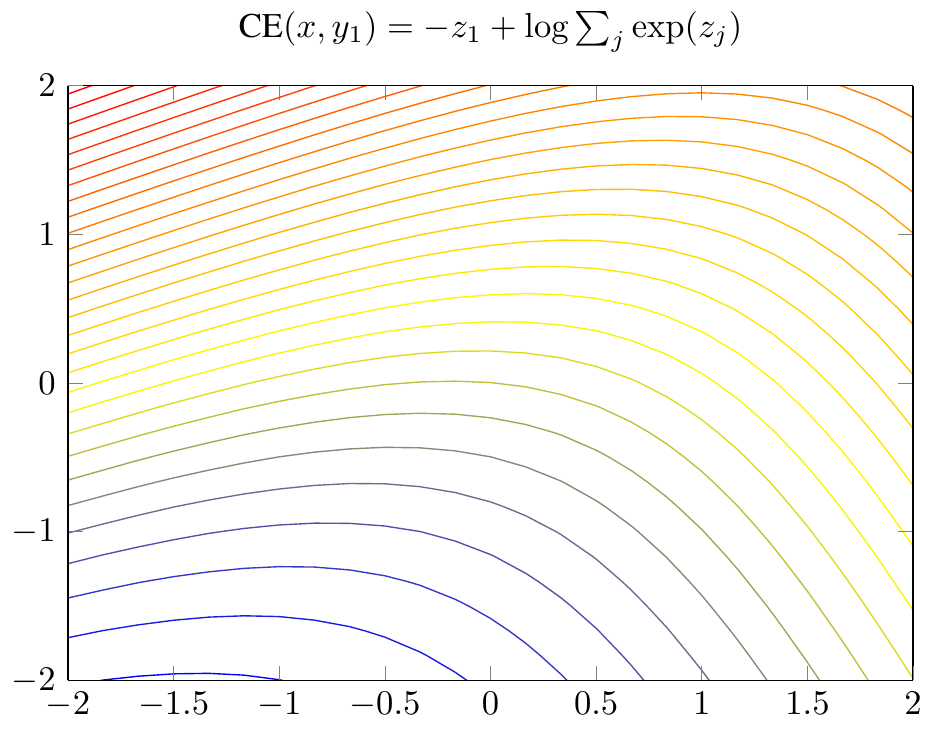}
    \includegraphics[width=0.45\textwidth]{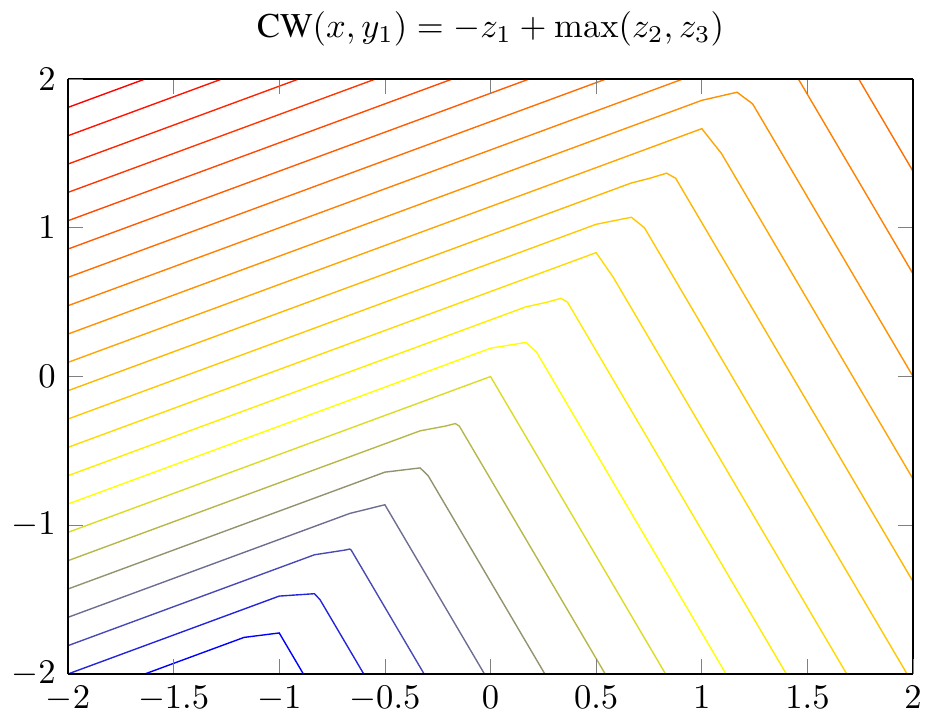}
    \includegraphics[width=0.45\textwidth]{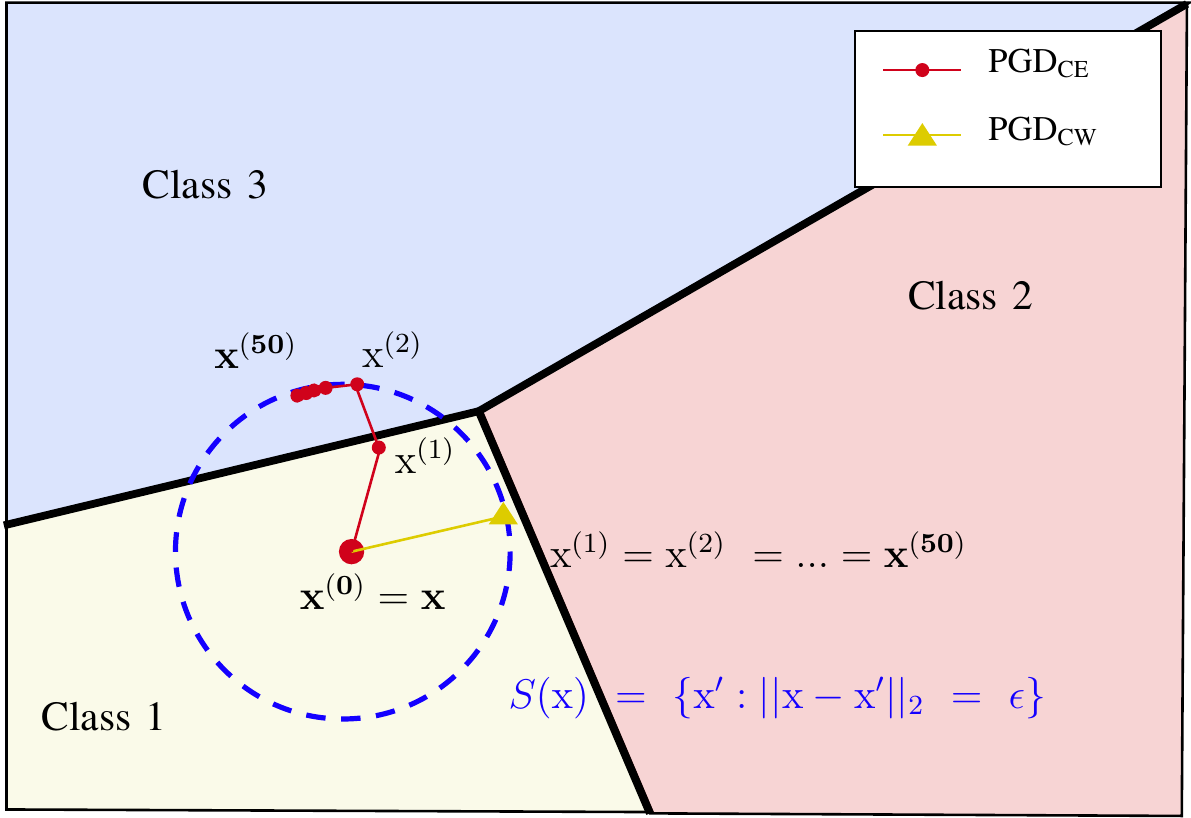} 
    \includegraphics[width=0.45\textwidth]{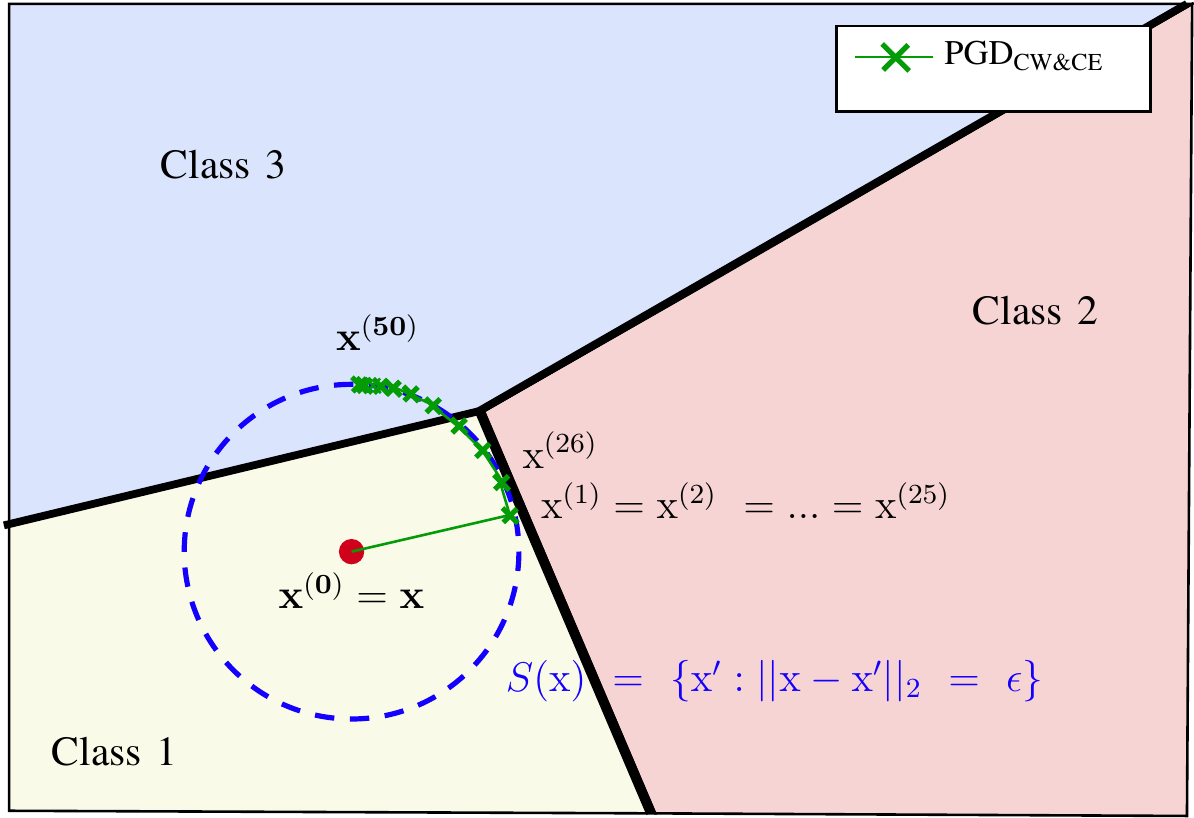} 
    \caption{\emph{Top row}: The level sets of CE and CW losses (w.r.t class $y=1$). \emph{Bottom row}: (Left) Intermediate PGD points, executed with a single surrogate, where red circles indicate PGD with CE and the yellow triangle PGD with CW, (Right) Intermediate PGD points, but here the objective changes in the middle point ($T=T/2$) of the procedure (green crosses). The blue dashed circle visualizes the boundary surface, which in this case is a disk of radius $\epsilon=0.4$ centered at $\vx$, of the feasible PGD solutions.}
    \label{fig:toy}
\end{figure*}

Our work is motivated by the observation that a single surrogate loss is unable to perform equally well across different robust models. Croce and Hein \cite{croce2020reliable} provide strong empirical evidence to back up this argument. Specifically, in their study they investigate the effectiveness of three objectives: CE, CW and DLR. These three aforementioned objectives have expressions that are distinguished by small differences, yet each option can profoundly influence the Attack Success Rate (ASR) of PGD. Of course, this phenomenon is not surprising at all: the optimization space coincides with the high-dimensional pixel space of natural images, hence even just a rescaling that links the CW with DLR loss is capable of producing non-trivial discrepancies in the respective loss landscapes. Above all, it is critical to bear in mind the surrogate loss as another hyperparameter, akin to step-size or optimizer, which has the potential of causing some degree of robustness overestimation on its own right.

The most straightforward mitigation for this behaviour is to aggregate many different formulations in the same run of PGD. The aggregation of objectives may be instantiated in a variety of ways. Our work is  based on a simple \emph{idea} for performing such an aggregation: Divide the PGD process into multiple successive stages, where the surrogate loss changes in the beginning of every stage, and the starting point of every stage coincides with the last step iterate of the previous one. This procedure, when using $K$ stages, can be formulated as:
$$ \cL(\vx,y) = 
    \begin{cases}
    \cL_1(\vx,y), & \text{if } t < \frac{T}{K} \\
    \cL_2(\vx,y), & \text{if } \frac{T}{K} \leq t < \frac{2T}{K} \\
    \hspace{0.5cm} \vdots \\
    \cL_K(\vx,y), & \text{if } \frac{(K-1)T}{K} \leq t < T 
    \end{cases}$$
In this paper, we will consider the cases where $K=2,3$, using for surrogates the most common choices: CE, CW and DLR. 

Notice how this alternation strategy can be viewed as a more complicated initialization: Each PGD stage starts from the initial point $\vx^{(0)} = \vx + \vect{\delta}$, where $\vect{\delta}$: the accumulated perturbation of all previous stages. Of course, an immediate extension is to consider variable starting timesteps $t_k$ for stage $k$, but in this work, we heuristically set equal time intervals between all stages. 

In the remaining discussion, our loss switching variant will be abbreviated as follows: $\text{PGD}_{\cL_1\&\cL_2\&...\&\cL_K}$, e.g., $\text{PGD}_{\text{CE}}$ for simple PGD with CE surrogate and $\text{PGD}_{\text{CE}\&\text{CW}}$ for two-stage PGD with CE and CW. \\

\begin{table*}[t]
    \centering
    \begin{tabular}{c c c c c c}
    \toprule 
    Dataset & \# & Paper & Model ID in \href{https://github.com/RobustBench/robustbench}{RobustBench leaderboard} & Architecture & Standard Acc. (\%) \\
    \midrule 
    \multirow{15}{*}{CIFAR-10}&1&\cite{robustness}             & Engstrom2019Robustness & ResNet-50 & 87.03\\
    &2&\cite{carmon2019unlabeled}    & Carmon2019Unlabeled    & WideResNet-28-10 & 89.69 \\
    &3&\cite{hendrycks2019using}     & Hendrycks2019Using     & WideResNet-28-10 & 87.11 \\
    &4&\cite{zhang2019you}           & Zhang2019You           & WideResNet-34-10 & 87.20  \\
    &5&\cite{zhang2019theoretically} & Zhang2019Theoretically & WideResNet-34-10 & 84.92 \\
    &6&\cite{wu2020adversarial}      & Wu2020Adversarial      & WideResNet-34-10 & 85.36 \\
    &7&\cite{sehwag2021robust}       & Sehwag2021Proxy\_R18   & ResNet-18 & 84.59\\
    &8&\cite{andriushchenko2020understanding} & Andriushchenko2020Understanding & PreActResNet-18  & 79.84 \\
    &9&\cite{dai2021parameterizing} & Dai2021Parameterizing                     & WideResNet-28-10 & 87.02 \\
    &10&\cite{gowal2021improving}   & Gowal2021Improving\_28\_10\_ddpm\_100m    & WideResNet-28-10 & 87.50\\
    &11&\cite{huang2021exploring}   & Huang2021Exploring\_ema    & WideResNet-34-R & 91.23 \\
    &12&\cite{zhang2020geometry}    & Zhang2020Geometry          & WideResNet-28-10 & 89.36 \\
    &13&\cite{rade2021helperbased}  & Rade2021Helper\_R18\_extra & PreActResNet-18 & 89.02 \\
    &14&\cite{addepalli2021towards} & Addepalli2021Towards\_RN18 & ResNet-18 & 80.24  \\
    &15&\cite{sehwag2020hydra}      & Sehwag2020Hydra            & WideResNet-28-10 & 88.98 \\ 
    \midrule 
    \multirow{6}{*}{CIFAR-100} & 1 & \cite{rade2021helperbased} & Rade2021Helper\_R18\_ddpm & PreActResNet-18 & 61.50\\
     & 2 & \cite{rebuffi2021fixing}     & Rebuffi2021Fixing\_R18\_ddpm & PreActResNet-18 & 56.87\\
     & 3 & \cite{addepalli2021towards}  & Addepalli2021Towards\_PARN18 & PreActResNet-18 & 62.02\\
     & 4 & \cite{Rice2020OverfittingIA} & Rice2020Overfitting          & PreActResNet-18 & 53.83\\
     & 5 & \cite{hendrycks2019using}    & Hendrycks2019Using           & WideResNet-28-10 & 59.23 \\
     & 6 & \cite{rebuffi2021fixing}     & Rebuffi2021Fixing\_28\_10\_cutmix\_ddpm & WideResNet-28-10 & 62.41 \\ 
    \midrule 
    \multirow{4}{*}{ImageNet} & 1 & \cite{Salman2019ProvablyRD} & Salman2020Do\_R18 & ResNet18 & 52.92 \\
      & 2 & \cite{Salman2020DoAR} & Salman2020Do\_R50      & ResNet50 & 64.02 \\
      & 3 & \cite{robustness}     & Engstrom2019Robustness & ResNet50 & 62.56 \\
      & 4 & \cite{Wong2020FastIB} & Wong2020Fast           & ResNet50 & 55.62 \\
    \bottomrule
    \end{tabular}
    \vspace{0.2cm}
    \caption{Our model collection, consisting of 25 $\linf$-bounded defenses obtained from the ModelZoo of RobustBench.}
    \label{table:correspondence}
\end{table*}

\section{Experiments}\label{sec:experiments}

\subsection{Toy Example.} 

We present a toy example which elucidates that using a single surrogate during PGD may deteriorate performance. Assume a 2D problem of 3-way classification (classes: $y_1,y_2,y_3$). Inputs are $\vx = (x_1,x_2)^T$ and the linear classifier is $\vz = (z_1,z_2,z_3)^T = \vect{W}\vx$, with:
$$ 
\vect{W} = \begin{bmatrix}
0.3 & -0.3  \\
1   & -0.01 \\
-0.25 & 0.75
\end{bmatrix}
$$
Consider an input $\vx = (-0.45,-0.8)$, belonging to the class $y_1$. The linear model classifies it correctly to its ground-truth class, since $z_1 > \max(z_2,z_3)$. Suppose that our goal is to generate a perturbation $\delta$ of bounded $\lt$-norm (say $\epsilon=0.4$). A straightforward way to achieve this is by executing PGD, maximizing a surrogate loss, e.g., CE or CW. For the input $\vx$ of class $y_1$, these losses are analytically calculated as:
\begin{equation*}
    \begin{split}
        \text{CE}(\vx,y) &= -z_1 + \log \big( \sum_{j=1}^3 \exp(z_j) \big) \\
        \text{CW}(\vx,y) &= -z_1 + \max(z_2,z_3)
    \end{split}
\end{equation*}
\autoref{fig:toy} illustrates the level sets of these two objectives. In the bottom left panel of \autoref{fig:toy}, we visualize the optimization trajectories of PGD for different choices of surrogates. The learning rate is held fixed to $\eta=2\epsilon$ and PGD is executed for $T=50$ iterations. The blue dashed circle denotes the boundary of the feasible region, whereas the circle, triangle and cross-shaped points show the intermediate points of PGD ($\vx^{(1)},...,\vx^{(50)}$). Using the CE as surrogate (red circle points) manages to successfully perturb the input $\vx$, but CW objective (yellow triangle points) fails because the linear level sets produce gradients that gets the optimization jammed on a single point. The bottom right panel, however, demonstrates that the loss alternation method (green cross points) isn't affected from the failure mode of CW and finds an adversary. 

Despite being restricted, this synthetic toy example underpins the argument that using multiple surrogates in the same run of PGD renders the overall procedure more ``robust" in the objective selection: Even if some individual choice is infertile for whatever reason, the other alternatives may be enough to find an adversary. 

\subsection{Models and Datasets}

We will conduct our experiments in a sizable collection of 25 $\linf$-bounded robust models. Specifically, the collection comprises of 15 and 6 defenses on CIFAR-10 and CIFAR-100 \cite{cifar10} respectively, trained with $\epsilon=8/255$, and 4 defenses on ImageNet \cite{imagenet}, trained with perturbation bound $\epsilon=4/255$. 
The models are pre-trained and readily obtained from the ModelZoo of RobustBench \cite{croce2021robustbench} library. Our collection's robust models originate from various recent works: \cite{robustness,carmon2019unlabeled,hendrycks2019using,zhang2019you,zhang2019theoretically,wu2020adversarial,sehwag2021robust,andriushchenko2020understanding,dai2021parameterizing,gowal2021improving,huang2021exploring,zhang2020geometry,rade2021helperbased,addepalli2021towards,sehwag2020hydra,Salman2020DoAR,rebuffi2021fixing,Rice2020OverfittingIA}. The architectures of these models are ResNets \cite{he2016deep} and Wide ResNets (WRN) \cite{zagoruyko2016wide}.
In \autoref{table:correspondence}, we exhibit our model collection: For each case (row), the classifier is matched with the respective paper/work, architecture, ModelID from RobustBench ModelZoo and the accuracy that the classifier attains on the respective clean evaluation set. In the case of CIFAR-10 and CIFAR-100, this coincides with the 10,000 images of the standard test set, whereas in the ImageNet case, 5,000 images from the val set are picked, accordingly to the established split of RobustBench library.  
We also state that in the following discussion, we'll refer to the terms Attack Success Rate (ASR) and Robust Accuracy (equal to $1-$ASR) interchangeably to quantify the strength of each attack. 

\subsection{Experimental Analysis}

\begin{table*}[t]
\centering
\resizebox{0.97\textwidth}{!}{
\begin{tabular}{crccccccc}
    \toprule
    & & \multicolumn{3}{c}{\emph{K=1}} & \multicolumn{3}{c}{\emph{K=2}} & \emph{K=3} \\
     \\ 
    & Model ID & $\ce$ & $\cw$ & $\dlr$ &  $\cecw$ & $\cedlr$ & $\cwdlr$ & $\cecwdlr$ \\ 
    \cmidrule(lr){2-2}  \cmidrule(lr){3-5} \cmidrule(lr){6-8} \cmidrule(lr){9-9}
    \parbox[t]{2mm}{\multirow{15}{*}{\rotatebox[origin=c]{90}{CIFAR-10}}} & Engstrom2019Robustness \cite{robustness}          & 52.24 & 52.59 & 53.55  & 50.29\fn{-1.95} & \textbf{50.22}{\fn{-2.02}}  & 52.63\fd{+0.04} & 50.27{\fn{-1.97}}           \\
    &Carmon2019Unlabeled \cite{carmon2019unlabeled}    & 62.09 & 60.86 & 61.16 & 60.00{\fn{-0.86}} & 60.00{\fn{-1.16}}          & 60.88\fd{+0.02} & \textbf{59.97}{\fn{-0.89}}\\
    &Hendrycks2019Using \cite{hendrycks2019using}      & 57.38 & 56.61 & 57.47 & 55.41{\fn{-1.20}} & 55.37{\fn{-2.01}}          & 56.55\fn{-0.06} & \textbf{55.35}\fn{-1.26} \\
    &Zhang2019You \cite{zhang2019you}                  & 46.28 & 47.44 & 47.97 & 45.33{\fn{-0.95}} & \textbf{45.32}{\fn{-0.96}} & 47.42\fn{-0.02} & \textbf{45.32}\fn{-0.96}  \\
    &Zhang2019Theoretically \cite{zhang2019theoretically} \dag & 55.47 & 54.21 & 54.39 & 53.45{\fn{-0.76}} & 53.43{\fn{-0.96}}     & 54.23\fd{+0.02} & \textbf{53.41}\fn{-0.80} \\
    &Wu2020Adversarial \cite{wu2020adversarial}         & 59.05 & 56.93 & 57.02 & 56.47{\fn{-0.46}} & 56.44{\fn{-0.58}}       & 56.94\fd{+0.01} & \textbf{56.42}\fn{-0.51}\\
    &Sehwag2021Proxy\_R18 \cite{sehwag2021robust}       & 58.68 & 57.22 & 57.89 & 56.06{\fn{-1.16}} & \textbf{56.05}{\fn{-1.84}} & 57.21\fd{-0.01} & 56.06\fn{-1.16}           \\
    &Andriushchenko2020Understanding \cite{andriushchenko2020understanding} & 47.14 & 46.62 & 47.62 & 44.56{\fn{-2.06}} & 44.53{\fn{-2.61}} & 46.62\fn{0} & \textbf{44.50}\fn{-2.12} \\
    &Dai2021Parameterizing \cite{dai2021parameterizing}  & 63.98 & 63.23 & 63.83 & 61.80{\fn{-1.43}} & \textbf{61.76}\fn{-2.07} & 63.23\fn{0} & 61.77\fn{-1.46}           \\
    &Gowal2021Improving\_28\_10\_ddpm\_100m \cite{gowal2021improving}     & 65.79 & 65.20 & 65.76 & 63.86{\fn{-1.34}} & 63.85\fn{-1.91} & 65.20\fn{0} & \textbf{63.84}\fn{-1.36}  \\
    &Huang2021Exploring\_ema \cite{huang2021exploring}        & 64.95 & 64.15 & 64.64 & 63.09{\fn{-1.06}} & \textbf{63.03}\fn{-1.61}    & 64.12\fn{-0.03} & 63.06\fn{-1.09}  \\
    &Zhang2020Geometry \cite{zhang2020geometry}               & 66.67 & 60.40 & 60.59 & 59.78{\fn{-0.62}} & \textbf{59.69}\fn{-0.90} & 60.37\fn{-0.03} & \textbf{59.69}\fn{-0.71} \\
    &Rade2021Helper\_R18\_extra \cite{rade2021helperbased}    & 61.48 & 58.51 & 58.56 & 57.77{\fn{-0.74}} & \textbf{57.74}\fn{-0.82} & 58.51\fn{0} & \textbf{57.74}\fn{-0.77}  \\
    &Addepalli2021Towards\_RN18 \cite{addepalli2021towards}   & 56.00 & 51.88 & 51.97 & 51.45{\fn{-0.43}} & 51.43\fn{-0.54}          & 51.86\fn{-0.03} & \textbf{51.41}\fn{-0.47}  \\
    &Sehwag2020Hydra \cite{sehwag2020hydra}                   & 59.86 & 58.41 & 58.57 & 57.66{\fn{-0.75}} & \textbf{57.61}\fn{-0.96} & 58.40\fn{-0.01} & \textbf{57.61}\fn{-0.80}  \\
    \midrule
    \parbox[t]{2mm}{\multirow{6}{*}{\rotatebox[origin=c]{90}{CIFAR-100}}}&Rade2021Helper\_R18\_ddpm~\cite{rade2021helperbased}    & 32.60 & 29.66 & 29.69  & 29.12\fn{-0.54} & 29.08\fn{-0.61}  & 29.66\fn{0}     & 29.12\fn{-0.54}  \\
    &Rebuffi2021Fixing\_R18\_ddpm~\cite{rebuffi2021fixing}            & 31.82 & 29.20 & 29.25  & 28.68\fn{-0.52} & \textbf{28.65}\fn{-0.60}  & 29.20\fn{0}     & 28.68\fn{-0.52}  \\
    &Addepalli2021Towards\_PARN18~\cite{addepalli2021towards}         & 32.90 & 28.10 & 28.23  & 27.68\fn{-0.42} & \textbf{27.63}\fn{-0.60}  & 28.10\fn{0}     & 27.67\fn{-0.43}  \\ 
    &Rice2020Overfitting~\cite{Rice2020OverfittingIA}                 & 20.89 & 20.42 & 20.62  & 19.33\fn{-1.09} & 19.33\fn{-1.29}  & 20.42\fn{0}     & \textbf{19.32}\fn{-1.10}  \\ 
    &Hendrycks2019Using~\cite{hendrycks2019using}                     & 33.17 & 30.84 & 32.34  & 29.43\fn{-1.41} & 29.43\fn{-2.91}  & 30.83\fn{-0.01} & \textbf{29.36}\fn{-1.48}  \\ 
    &Rebuffi2021Fixing\_28\_10\_cutmix\_ddpm~\cite{rebuffi2021fixing} & 35.74 & 33.60 & 33.67 & 32.53\fn{-1.07} & \textbf{32.50}\fn{-1.17} & 33.60\fn{0} & 32.53\fn{-1.07} \\  
    \midrule 
    \parbox[t]{2mm}{\multirow{4}{*}{\rotatebox[origin=c]{90}{ImageNet}}} &Salman2020Do\_R18~\cite{Salman2020DoAR}      & 29.50 & 27.32 & 27.60 & 25.64\fn{-1.68} & 2\textbf{5.62}\fn{-1.98} & 27.32\fn{0} & 25.66\fn{-1.66}    \\  
    &Salman2020Do\_R50~\cite{Salman2020DoAR}      & 38.78 & 37.62 & 38.04 & 35.30\fn{-2.32} & \textbf{35.26}\fn{-2.78} & 37.64\fd{+0.02}  & \textbf{35.26}\fn{-2.36}    \\  
    &Engstrom2019Robustness~\cite{robustness}     & 32.64 & 32.64 & 33.16 & 30.00\fn{-2.64} & \textbf{29.94}\fn{-2.70} & 32.66\fd{+0.02}  & 29.96\fn{-2.68}    \\  
    &Wong2020Fast~\cite{Wong2020FastIB}           & 27.50 & 27.46 & 27.86 & 25.76\fn{-1.70} & \textbf{25.74}\fn{-1.72}  & 27.48\fd{+0.02} & \textbf{25.74}\fn{-1.72}    \\  
    \bottomrule
    \end{tabular}
}
\vspace{0.2cm}
\caption{Comparing single-loss PGD with the multi-stage variant of PGD (with $K=2,3$). PGD starts from the clean point (no added noise). The experiments are executed for $T=100$ with no restarts. Each entry reports the robust accuracy of each classifier for the given method. (\dag): Attacked with $\epsilon=0.031$. The \textcolor{dgreen}{green} (\textcolor{dred}{red}) numbers indicate the relative decrease (increase) of robust accuracy with respect to the best single-loss attack of each multi-loss variant.} 
\label{table:t100its}
\end{table*}

\subsubsection{Multi-Stage PGD versus Single-Loss}
First, we compare the loss alternation strategy against the typical single loss variants of PGD. In this experimental setting, step size is held fixed to $\eta^{(t)}=\epsilon/4$ and the optimizer is set to standard gradient with the sign operation. Our computational budget is $T=100$ with no restarts. Since no restarts are used, we choose to initiate PGD from the clean points (no initial perturbation) in order to eliminate any source of randomness in the results. \autoref{table:t100its} presents the robust accuracy obtained of PGD with different choices of surrogates, for every classifier in our collection.

Overall, there are several noteworthy remarks: First, the single-loss columns ($K=1$) demonstrate that the surrogate loss can greatly affect the ASR of PGD, confirming the findings of previous studies, as that of Croce and Hein \cite{croce2020reliable}.  On average, margin loss is the most reliable option but there are cases where it performs worse than CE. There are instances where CE lags behind the other two options by a large margin, e.g., as in the model from \cite{addepalli2021towards} (\texttt{Addepalli2021Towards\_RN18}), where the gap is greater than $4\%$. This indicates that it is impossible to select \emph{a priori} the best possible objective for a given model. This observation consitutes strong evidence that no surrogate loss is reliable enough on its own.

Next, the results highlight the advantage of using multiple losses in the same run of PGD: When combining CE with CW or DLR ($\cecw$ and $\cedlr$ columns), or both ($\cecwdlr$ column) the attack is always stronger (lower rob. acc.) than the respective single-loss PGD. 
On average, $\cecw$ and $\cedlr$ decrease robust accuracy by 1.05\% and 1.39\% (absolute) respectively, over their corresponding single-loss variants in the CIFAR-10 case. In CIFAR-100, the average absolute decrease in the robust accuracy of the models is 0.81\% and 1.19\% for $\cecw$ and $\cedlr$. For the ImageNet dataset, the alternation strategy provides even greater improvements, since the corresponding average reduction reaches 2.08\% and 2.29\%
In the case of $\cwdlr$, the obtained ASR is nearly identical with $\cw$, implying that the alternation step in this case may be futile due to the similarity between the expressions of CW and DLR losses. Overall, our experiments illustrate that the alternation strategy is highly beneficial, across all models and datasets. 

\begin{table*}[t]
\centering
\begin{tabular}{r  c c c >{\bfseries}c c }
    \toprule
    Model ID & $\apgdce$ & $\apgddlr$ & FAB & \normalfont{$\cecwdlr$} & $\Delta$ \\ 
    \cmidrule(lr){1-1}  \cmidrule(lr){2-2} \cmidrule(lr){3-3} \cmidrule(lr){4-4} \cmidrule(lr){5-5} \cmidrule(lr){6-6} 
    Engstrom2019Robustness \cite{robustness}       & 51.72 & 52.67 & 50.67 & 50.27 & -0.40 \\
    Carmon2019Unlabeled \cite{carmon2019unlabeled} & 61.74 & 60.67 & 60.88 & 59.97 & -0.70 \\
    Hendrycks2019Using \cite{hendrycks2019using}  & 57.23 & 57.03 & 55.55 & 55.35  & -0.20 \\
    Zhang2019You \cite{zhang2019you}              & 46.15 & 47.39 & 45.83 & 45.32  & -0.51 \\
    Zhang2019Theoretically \cite{zhang2019theoretically} \dag      & 55.28 & 53.52 & 53.92 & 53.41 & -0.11 \\
    Wu2020Adversarial \cite{wu2020adversarial}  & 58.90 & 56.68 & 56.82 & 56.42 & -0.26 \\
    Sehwag2021Proxy\_R18\cite{sehwag2021robust} & 58.38 & 57.37 & 56.27 & 56.06 & -0.21 \\
    Andriushchenko2020Understanding \cite{andriushchenko2020understanding}  & 46.93 & 47.08 & 44.72 & 44.50 & -0.22 \\
    Dai2021Parameterizing \cite{dai2021parameterizing}               & 63.93 & 63.44 & 62.27 & 61.77 & -0.50 \\
    Gowal2021Improving\_28\_10\_ddpm\_100m\cite{gowal2021improving}  & 65.63 & 65.14 & 64.14 & 63.84 & -0.30 \\
    Huang2021Exploring\_ema \cite{huang2021exploring}        & 64.55 & 64.14 & 64.45 & 63.06 & -1.08\\
    Zhang2020Geometry \cite{zhang2020geometry}               & 66.37 & 60.19 & 59.97 & 59.69 & -0.28\\
    Rade2021Helper\_R18\_extra \cite{rade2021helperbased}    & 61.40 & 58.41 & 58.42 & 57.74 & -0.67\\
    Addepalli2021Towards\_RN18 \cite{addepalli2021towards}   & 55.80 & 51.56 & 51.93 & 51.41 & -0.15\\
    Sehwag2020Hydra \cite{sehwag2020hydra}                   & 59.60 & 58.29 & 58.29 & 57.61 & -0.68 \\
    \bottomrule
    \end{tabular}
\vspace{0.2cm}
\caption{Comparing $\cecwdlr$ with the untargeted version of every single white-box component from the AutoAttack ensemble.  Each entry reports the robust accuracy of each classifier for the given method.  $\Delta$ column report the robust accuracy gap between $\cecwdlr$ and the best among the AutoAttack components. The experiments are executed for $T=100$ with no restarts. (\dag): Attacked with $\epsilon=0.031$.}
\label{table:autoattack}
\end{table*}

Finally, it is illustrated that on average $\cecwdlr$ is better than $\cecw$ and $\cedlr$ (mainly on the CIFAR-10 case), yet the differences are small. In some cases, using the alternation scheme with two stages is better than $\cecwdlr$. This informs us that it is not always better to add another stage/objective in the alternation process. In a fixed iteration budget, adding another loss reduces the overall time allotted to each stage. We assume that this hurts performance because the reduced number of iterations is not enough to reach the stagnating region of each loss.  \\ 

\begin{table*}[!ht]
\centering
\hspace*{-1cm}
\resizebox{0.97\textwidth}{!}{
\begin{tabular}{r c c c |  c c c}
    \toprule
    Model ID & GAMA-PGD \cite{Sriramanan2020GuidedAA} & $\cecwdlr$   & $\Delta$  &  MD Attack \cite{Ma2020ImbalancedGA} & $\cecwdlr$  & $\Delta$\\
          & &  (GAMA-PGD sch.) & & & (MD sched.) &    \\
    \midrule
    Engstrom2019Robustness~\cite{robustness}       & 50.05 & \textbf{49.88}         & -0.17 & 50.34 & \textbf{49.87} & -0.47\\
    Carmon2019Unlabeled~\cite{carmon2019unlabeled} & 59.84 & \textbf{59.78}         & -0.06 & 59.83 & \textbf{59.72} & -0.11\\
    Hendrycks2019Using~\cite{hendrycks2019using}   & \textbf{55.22} & 55.26         & +0.04 & \textbf{55.15} & 55.20 & +0.05\\
    Zhang2019You~\cite{zhang2019you}               & 45.32 & \textbf{45.20}         & -0.12 & 45.49 & \textbf{45.17} & -0.32\\
    Zhang2019Theoretically~\cite{zhang2019theoretically}\dag & 53.29          & 53.29           & 0      & 53.36 & \textbf{53.26} & -0.10\\
    Wu2020Adversarial~\cite{wu2020adversarial}               & 56.30          & 56.30           & 0      & 56.28 & \textbf{56.26} & -0.02\\
    Sehwag2021Proxy\_R18~\cite{sehwag2021robust}            &  56.01          & \textbf{55.95}  & -0.06  & 55.92 & \textbf{55.89} & -0.03\\
    Andriushchenko2020Understanding~\cite{andriushchenko2020understanding}  & 44.42 & \textbf{44.41}   & -0.01 & 44.57 & \textbf{44.44} & -0.13\\
    Dai2021Parameterizing~\cite{dai2021parameterizing}                & 61.94 &\textbf{61.74}          & -0.20 & 61.99 & \textbf{61.72}   & -0.27 \\
    Gowal2021Improving\_28\_10\_ddpm\_100m~\cite{gowal2021improving}  & 63.78 & \textbf{63.72}& -0.06  & 63.94 &  \textbf{63.73} & -0.21\\
    Huang2021Exploring\_ema~\cite{huang2021exploring}                 & \textbf{62.87} & 62.89 & +0.02 & 62.93 &  \textbf{62.86} & -0.07\\
    Zhang2020Geometry~\cite{zhang2020geometry}                        & 60.72 & \textbf{59.62}         & -1.10 & 59.73 & \textbf{59.58} & -0.15\\
    Rade2021Helper\_R18\_extra~\cite{rade2021helperbased}             & 57.78 & \textbf{57.73}         & -0.05 & 57.74 & \textbf{57.72} & -0.02\\
    Addepalli2021Towards\_RN18~\cite{addepalli2021towards}            & 51.43 & \textbf{51.26}         & -0.17 & 51.30 &  \textbf{51.25}& -0.05\\
    Sehwag2020Hydra~\cite{sehwag2020hydra}                            & 57.49 & \textbf{57.43}         & -0.06 & \textbf{57.31} & 57.45 & +0.14\\
    \bottomrule
\end{tabular}
}
\vspace{0.2cm}
\caption{Comparing $\cecwdlr$ with the strongest attacks of our computational budget ($\mathbf{T=100, R=1}$). Each entry reports the robust accuracy of each classifier for the given method. $\Delta$ columns report the robust accuracy gap between the compared methods.  (\dag): Attacked with $\epsilon=0.031$. }
\label{table:baselines}
\end{table*}
\begin{table*}[!ht]
\centering
\hspace*{-1cm}
\resizebox{0.97\textwidth}{!}{
\begin{tabular}{r c c c |  c c c}
    \toprule
    Model ID & GAMA-PGD \cite{Sriramanan2020GuidedAA} & $\cecwdlr$   & $\Delta$  &  MD Attack \cite{Ma2020ImbalancedGA} & $\cecwdlr$  & $\Delta$\\
          & &  (GAMA-PGD sch.) & & & (MD sched.) &    \\
    \midrule
    Engstrom2019Robustness~\cite{robustness}                 & 49.88 & \textbf{49.80}   & -0.08  & 50.13 & \textbf{49.68} & -0.45 \\
    Carmon2019Unlabeled~\cite{carmon2019unlabeled}           & \textbf{59.70} & 59.71   & +0.01  & 59.67 & 59.67          & 0     \\
    Hendrycks2019Using~\cite{hendrycks2019using}             & 55.21 & \textbf{55.09}   & -0.12  & \textbf{55.10} & 55.13 & +0.03 \\
    Zhang2019You~\cite{zhang2019you}                         & 45.02 & 45.02            & 0      & 45.36 & \textbf{45.00} & -0.36 \\
    Zhang2019Theoretically~\cite{zhang2019theoretically}\dag & 53.26 & \textbf{53.20}   & -0.06  & 53.26 & \textbf{53.19} & -0.07 \\
    Wu2020Adversarial~\cite{wu2020adversarial}               & 56.29 & \textbf{56.23}   & -0.06  & 56.24 & \textbf{56.20} & -0.04 \\
    Sehwag2021Proxy\_R18~\cite{sehwag2021robust}             & 55.89 & \textbf{55.85}   & -0.04  & 55.85 & \textbf{55.81} & -0.04 \\
    Andriushchenko2020Understanding~\cite{andriushchenko2020understanding}& 44.32 & \textbf{44.29} & -0.03 & 44.42 & \textbf{44.36} & -0.06  \\
    Dai2021Parameterizing~\cite{dai2021parameterizing}                & 61.95 & \textbf{61.70}  & -0.25 & 61.83 & \textbf{61.65}  & -0.18 \\
    Gowal2021Improving\_28\_10\_ddpm\_100m~\cite{gowal2021improving}  & 63.81 & \textbf{63.66}  & -0.15  & 63.90 & \textbf{63.65} & -0.25 \\
    Huang2021Exploring\_ema~\cite{huang2021exploring}                 & 62.77 & \textbf{62.74}  & -0.03  & 62.80 & \textbf{62.71} & -0.09 \\
    Zhang2020Geometry~\cite{zhang2020geometry}                        & 60.28 & \textbf{59.44}  & -0.88  & 59.59 & \textbf{59.50} & -0.09 \\
    Rade2021Helper\_R18\_extra~\cite{rade2021helperbased}             & 57.74 & \textbf{57.66}  & -0.08  & 57.71 & \textbf{57.67} & -0.04 \\
    Addepalli2021Towards\_RN18~\cite{addepalli2021towards}            & 51.46 & \textbf{51.23}  & -0.23  & 51.22 & \textbf{51.21} & -0.01 \\
    Sehwag2020Hydra~\cite{sehwag2020hydra}                            & 57.37 & \textbf{57.31}  & -0.06  & \textbf{57.23} & 57.34 & +0.11 \\
    \bottomrule
\end{tabular}
}
\vspace{0.2cm}
\caption{Comparing $\cecwdlr$ with the similar (to our work) baselines ($\mathbf{T=100, R=5}$). Each entry reports the robust accuracy of each classifier for the given method. $\Delta$ columns report the robust accuracy gap between the compared methods.  (\dag): Attacked with $\epsilon=0.031$. }
\label{table:baselines_v2}
\end{table*}

For the remainder of the experimental section, we will focus on the adversarial defenses of CIFAR-10 dataset.

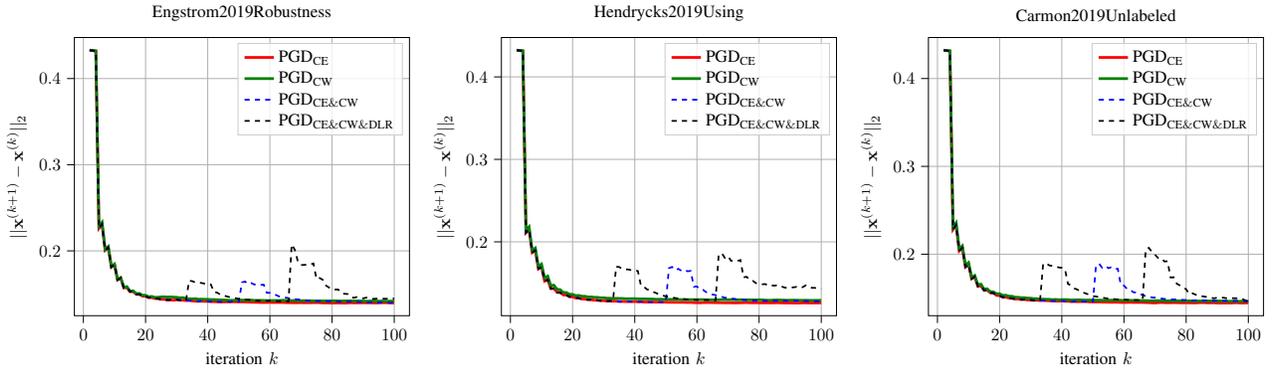
\begin{figure*}
    \centering
    \scalebox{.65}{
\begin{tikzpicture}

\definecolor{darkgray176}{RGB}{176,176,176}
\definecolor{green}{RGB}{0,128,0}
\definecolor{lightgray204}{RGB}{204,204,204}

\begin{axis}[
legend cell align={left},
legend style={fill opacity=0.8, draw opacity=1, text opacity=1, draw=lightgray204},
tick align=outside,
tick pos=left,
title={Engstrom2019Robustness},
x grid style={darkgray176},
xlabel={iteration \(\displaystyle k\)},
xmajorgrids,
xmin=-2.9, xmax=104.9,
xtick style={color=black},
y grid style={darkgray176},
ylabel={\(\displaystyle ||\mathbf{x}^{(k+1)} - \mathbf{x}^{(k)}||_2\)},
ymajorgrids,
ymin=0.124766998887062, ymax=0.447634990215302,
ytick style={color=black}
]
\addplot [ultra thick, red]
table {%
2 0.432945640981197
3 0.432465310692787
4 0.432069671154022
5 0.226150951981545
6 0.230911965668201
7 0.200687635987997
8 0.203992340862751
9 0.181157164275646
10 0.184059852361679
11 0.166353045403957
12 0.168672979176044
13 0.157022698223591
14 0.157483656331897
15 0.15257669441402
16 0.152918962240219
17 0.149566632434726
18 0.149587948545814
19 0.14748208142817
20 0.147360124960542
21 0.145669825077057
22 0.145845594257116
23 0.144629189223051
24 0.144567649066448
25 0.143724536597729
26 0.143736463114619
27 0.142993658781052
28 0.143140376284719
29 0.142665241956711
30 0.143308052644134
31 0.142865252494812
32 0.142716067954898
33 0.1423094381392
34 0.142433211579919
35 0.141942770034075
36 0.141624319776893
37 0.141445457935333
38 0.141660936027765
39 0.141354572027922
40 0.141417719423771
41 0.141031339690089
42 0.141114406883717
43 0.140877564549446
44 0.140759134441614
45 0.140735773593187
46 0.140869743004441
47 0.14083353087306
48 0.140929503887892
49 0.14085542306304
50 0.140819412693381
51 0.140768023654819
52 0.140792111828923
53 0.140582896471024
54 0.140407459288836
55 0.140326166450977
56 0.140303395912051
57 0.140069758072495
58 0.140106177330017
59 0.140119204893708
60 0.140230794548988
61 0.140044582858682
62 0.140127909705043
63 0.140117993652821
64 0.140114905536175
65 0.139957517012954
66 0.1399065836519
67 0.139847051501274
68 0.139811387583613
69 0.139815078228712
70 0.139943644031882
71 0.139764032363892
72 0.139861221686006
73 0.13983638599515
74 0.139929504022002
75 0.13994939647615
76 0.139876210987568
77 0.139626708477736
78 0.139681689962745
79 0.139598438367248
80 0.139496019035578
81 0.139529571682215
82 0.13961270481348
83 0.139513597786427
84 0.139442816674709
85 0.139545826539397
86 0.139817989319563
87 0.139697004631162
88 0.139676191359758
89 0.139723834469914
90 0.139792984798551
91 0.13969853259623
92 0.139736037924886
93 0.139639170467854
94 0.139529001042247
95 0.139678847268224
96 0.139823383763433
97 0.139892233237624
98 0.139756172299385
99 0.139593083560467
100 0.139584460556507
};
\addlegendentry{$\ce$}
\addplot [ultra thick, green]
table {%
2 0.432959172427654
3 0.432523975372314
4 0.432134045362473
5 0.228241047412157
6 0.233141741752625
7 0.201528981328011
8 0.205008928477764
9 0.182117879539728
10 0.184663839638233
11 0.166720733940601
12 0.168747822493315
13 0.157974053323269
14 0.158515521213412
15 0.153362405821681
16 0.153622894287109
17 0.150369024276733
18 0.150647231042385
19 0.148455450162292
20 0.148499653041363
21 0.146739596948028
22 0.147127165645361
23 0.146042231544852
24 0.14615295521915
25 0.146522802934051
26 0.146688503399491
27 0.146333930045366
28 0.146537961959839
29 0.146249928325415
30 0.1461121622473
31 0.145517560169101
32 0.14576038107276
33 0.144681912660599
34 0.144445114955306
35 0.144251095876098
36 0.144554426372051
37 0.143798302561045
38 0.14392758525908
39 0.143778650984168
40 0.143769859448075
41 0.14348821491003
42 0.143468203321099
43 0.143274521231651
44 0.143098056986928
45 0.142953955680132
46 0.142881691530347
47 0.142543400377035
48 0.142587761580944
49 0.142567206919193
50 0.14272715985775
51 0.142798581048846
52 0.142934013828635
53 0.142795282676816
54 0.142849781215191
55 0.142514991462231
56 0.142518959119916
57 0.142529238685966
58 0.142498462125659
59 0.14232224136591
60 0.142333628311753
61 0.142387413829565
62 0.142411032915115
63 0.142403607219458
64 0.14263110242784
65 0.142607281059027
66 0.142464973032475
67 0.142279240265489
68 0.142470763251185
69 0.14214902497828
70 0.142000223994255
71 0.142204075828195
72 0.142119896560907
73 0.14201677672565
74 0.141967056319118
75 0.141940680146217
76 0.141929770261049
77 0.141899413019419
78 0.141888834163547
79 0.14179993018508
80 0.142000821307302
81 0.142043074592948
82 0.141939853280783
83 0.141755173802376
84 0.141904371455312
85 0.141987076699734
86 0.142002101317048
87 0.141967835724354
88 0.141924244388938
89 0.141934558674693
90 0.141762471422553
91 0.141586468890309
92 0.141722362190485
93 0.14187032930553
94 0.141790700256824
95 0.141614484041929
96 0.141767088398337
97 0.141671956926584
98 0.141529188752174
99 0.141650077998638
100 0.141838546097279
};
\addlegendentry{$\cw$}
\addplot [line width=1pt, blue, dashed]
table {%
2 0.432945640981197
3 0.432465310692787
4 0.432069671154022
5 0.226150951981545
6 0.230911965668201
7 0.200687635987997
8 0.203992340862751
9 0.181157164275646
10 0.184059852361679
11 0.166353045403957
12 0.168672979176044
13 0.157022698223591
14 0.157483656331897
15 0.15257669441402
16 0.152918962240219
17 0.149566632434726
18 0.149587948545814
19 0.14748208142817
20 0.147360124960542
21 0.145669825077057
22 0.145845594257116
23 0.144629189223051
24 0.144567649066448
25 0.143724536597729
26 0.143736463114619
27 0.142993658781052
28 0.143140376284719
29 0.142665241956711
30 0.143308052644134
31 0.142865252494812
32 0.142716067954898
33 0.1423094381392
34 0.142433211579919
35 0.141942770034075
36 0.141624319776893
37 0.141445457935333
38 0.141660936027765
39 0.141354572027922
40 0.141417719423771
41 0.141031339690089
42 0.141114406883717
43 0.140877564549446
44 0.140759134441614
45 0.140735773593187
46 0.140869743004441
47 0.14083353087306
48 0.140929503887892
49 0.14085542306304
50 0.140819412693381
51 0.163366550281644
52 0.164424511194229
53 0.161378115341067
54 0.162612750008702
55 0.160885472893715
56 0.161609874516726
57 0.160455403104424
58 0.160783954784274
59 0.151347922533751
60 0.151232125163078
61 0.149048536047339
62 0.148761285468936
63 0.147092837914824
64 0.146816054359078
65 0.1458269854635
66 0.145139270797372
67 0.144157705977559
68 0.144220080599189
69 0.143497552201152
70 0.143457603231072
71 0.142608294561505
72 0.142420341521502
73 0.142273117154837
74 0.142244345545769
75 0.141872588470578
76 0.141680290922523
77 0.14151708304882
78 0.141632567197084
79 0.1415235748142
80 0.141513851359487
81 0.141067065000534
82 0.141069290712476
83 0.141075858771801
84 0.141192258670926
85 0.140919913575053
86 0.140908064246178
87 0.140710176154971
88 0.140687391236424
89 0.140469562113285
90 0.140289721935987
91 0.140152817443013
92 0.140262100547552
93 0.140143794715405
94 0.140158421397209
95 0.140050901398063
96 0.140000415742397
97 0.139963760226965
98 0.139929903745651
99 0.139670135378838
100 0.139690760597587
};
\addlegendentry{$\cecw$}
\addplot [line width=1pt, black, dashed]
table {%
2 0.432945640981197
3 0.432465310692787
4 0.432069671154022
5 0.226150951981545
6 0.230911965668201
7 0.200687635987997
8 0.203992340862751
9 0.181157164275646
10 0.184059852361679
11 0.166353045403957
12 0.168672979176044
13 0.157022698223591
14 0.157483656331897
15 0.15257669441402
16 0.152918962240219
17 0.149566632434726
18 0.149587948545814
19 0.14748208142817
20 0.147360124960542
21 0.145669825077057
22 0.145845594257116
23 0.144629189223051
24 0.144567649066448
25 0.143724536597729
26 0.143736463114619
27 0.142993658781052
28 0.143140376284719
29 0.142665241956711
30 0.143308052644134
31 0.142865252494812
32 0.142716067954898
33 0.1423094381392
34 0.165298659428954
35 0.164891069531441
36 0.163265920057893
37 0.163064377829432
38 0.162747851461172
39 0.161967240199447
40 0.161936942860484
41 0.160997807383537
42 0.152219038233161
43 0.151577295511961
44 0.149381129965186
45 0.148505974933505
46 0.147255296483636
47 0.146379738077521
48 0.145726420804858
49 0.145044459402561
50 0.144198883771896
51 0.143828040063381
52 0.143575210645795
53 0.143210151940584
54 0.142858514040709
55 0.142502871751785
56 0.142243329063058
57 0.142081213071942
58 0.141965428814292
59 0.142092969119549
60 0.14216547973454
61 0.14196551784873
62 0.1417954351753
63 0.141563034579158
64 0.141416486278176
65 0.141474928483367
66 0.141471517533064
67 0.207483603358269
68 0.201401144638658
69 0.187793781086802
70 0.183462869748473
71 0.18378656513989
72 0.183801792189479
73 0.184176242947578
74 0.185326816365123
75 0.167293043732643
76 0.168481954932213
77 0.166488341838121
78 0.159879877641797
79 0.157466803640127
80 0.157089103907347
81 0.155164319574833
82 0.155093149468303
83 0.150846693441272
84 0.150157950296998
85 0.149930756315589
86 0.146395429745316
87 0.145529632121325
88 0.146409694775939
89 0.145696442499757
90 0.144849658533931
91 0.145698046013713
92 0.145580998808146
93 0.144404874742031
94 0.144425347223878
95 0.144849410951138
96 0.144506424590945
97 0.144403950273991
98 0.144424333125353
99 0.143689769133925
100 0.144928935468197
};
\addlegendentry{$\cecwdlr$}
\end{axis}

\end{tikzpicture}}
    \scalebox{.65}{
\begin{tikzpicture}

\definecolor{darkgray176}{RGB}{176,176,176}
\definecolor{green}{RGB}{0,128,0}
\definecolor{lightgray204}{RGB}{204,204,204}

\begin{axis}[
legend cell align={left},
legend style={fill opacity=0.8, draw opacity=1, text opacity=1, draw=lightgray204},
tick align=outside,
tick pos=left,
title={Hendrycks2019Using},
x grid style={darkgray176},
xlabel={iteration \(\displaystyle k\)},
xmajorgrids,
xmin=-2.9, xmax=104.9,
xtick style={color=black},
y grid style={darkgray176},
ylabel={\(\displaystyle ||\mathbf{x}^{(k+1)} - \mathbf{x}^{(k)}||_2\)},
ymajorgrids,
ymin=0.110454904675484, ymax=0.44773694729805,
ytick style={color=black}
]
\addplot [ultra thick, red]
table {%
2 0.432405945360661
3 0.431935333609581
4 0.431613813340664
5 0.211112053543329
6 0.215224428772926
7 0.186940757632256
8 0.189629931300879
9 0.167243120372295
10 0.169617899358273
11 0.152320073768497
12 0.154130630940199
13 0.143220230266452
14 0.14355772472918
15 0.138699081018567
16 0.138654865548015
17 0.135611926317215
18 0.135551123544574
19 0.133404959440231
20 0.133297688364983
21 0.131910668835044
22 0.131790238320827
23 0.130623718574643
24 0.130634729042649
25 0.129775781333446
26 0.129781458824873
27 0.129171403646469
28 0.129158864244819
29 0.128612921833992
30 0.128683562651277
31 0.128188294172287
32 0.128267593309283
33 0.128069734796882
34 0.128087114542723
35 0.127608206868172
36 0.12773066021502
37 0.127541430070996
38 0.127543078735471
39 0.127449921295047
40 0.127592019587755
41 0.127291461974382
42 0.127359687760472
43 0.127012667432427
44 0.127028817161918
45 0.126942509710789
46 0.126961007416248
47 0.126851021125913
48 0.126905018985271
49 0.126835648939013
50 0.12693325355649
51 0.126688082739711
52 0.126758739650249
53 0.126657389178872
54 0.126618243232369
55 0.126648921519518
56 0.126606977805495
57 0.126439201384783
58 0.126583388894796
59 0.126253441497684
60 0.126195700094104
61 0.126085363477468
62 0.126337687298656
63 0.126417044773698
64 0.126385567560792
65 0.126317736059427
66 0.126279327347875
67 0.126163717955351
68 0.126170993149281
69 0.126057583987713
70 0.126102052927017
71 0.126048884093761
72 0.125990779697895
73 0.126004796400666
74 0.126096181049943
75 0.126252949908376
76 0.126382584795356
77 0.126128591746092
78 0.126088481992483
79 0.12613791577518
80 0.126096291765571
81 0.125984027385712
82 0.126024839282036
83 0.126000506207347
84 0.125963994339108
85 0.126047132909298
86 0.126144005581737
87 0.125994598418474
88 0.125928380265832
89 0.125923245698214
90 0.125883215218782
91 0.125797837749124
92 0.125785906612873
93 0.125789436027408
94 0.126005246266723
95 0.126009756028652
96 0.125922928005457
97 0.12587652310729
98 0.125884978920221
99 0.125942432582378
100 0.126008784174919
};
\addlegendentry{$\ce$}
\addplot [ultra thick, green]
table {%
2 0.432367626130581
3 0.431892134845257
4 0.431529140472412
5 0.21418309673667
6 0.218275561034679
7 0.189723498672247
8 0.192501925379038
9 0.170313284024596
10 0.173017280995846
11 0.156446332111955
12 0.158416630104184
13 0.146726559624076
14 0.147448365986347
15 0.142270499169827
16 0.142067553028464
17 0.138553541898727
18 0.138554762229323
19 0.136394852697849
20 0.136294859573245
21 0.134811728298664
22 0.135005874559283
23 0.133984527438879
24 0.13400567419827
25 0.13311060577631
26 0.133244854062796
27 0.132682534828782
28 0.13268411770463
29 0.132136490941048
30 0.132163519337773
31 0.131641071587801
32 0.131591708660126
33 0.131308257430792
34 0.131407746896148
35 0.131342454180121
36 0.13133707433939
37 0.131001540124416
38 0.131023009493947
39 0.131053770333529
40 0.131103638708591
41 0.130798152983189
42 0.130797758549452
43 0.130775233507156
44 0.130755088701844
45 0.130607357472181
46 0.130604491382837
47 0.130284781232476
48 0.130384810343385
49 0.13023789703846
50 0.13029663041234
51 0.13019015006721
52 0.130139025598764
53 0.130125121623278
54 0.130123054236174
55 0.129946916624904
56 0.1299210434407
57 0.129895602241158
58 0.129941455721855
59 0.129911671429873
60 0.129879060983658
61 0.129923136234283
62 0.129963571503758
63 0.129811240211129
64 0.129653717055917
65 0.129633573666215
66 0.1296924405545
67 0.129717059060931
68 0.129990300014615
69 0.129885305389762
70 0.129830272123218
71 0.129852437078953
72 0.129821497052908
73 0.129626380726695
74 0.129840169623494
75 0.12972289763391
76 0.129633860215545
77 0.129686505049467
78 0.12961874589324
79 0.129597984105349
80 0.129730276614428
81 0.129561653584242
82 0.129317448735237
83 0.129123006910086
84 0.129212855175138
85 0.129441733062267
86 0.129525176212192
87 0.129368307888508
88 0.129294716343284
89 0.12914230145514
90 0.129236890226603
91 0.129271042123437
92 0.129252031967044
93 0.129081171900034
94 0.129099077582359
95 0.129154810830951
96 0.129063839837909
97 0.128967345282435
98 0.129092652350664
99 0.129093348905444
100 0.129055338129401
};
\addlegendentry{$\cw$}
\addplot [line width=1pt, blue, dashed]
table {%
2 0.432405945360661
3 0.431935333609581
4 0.431613813340664
5 0.211112053543329
6 0.215224428772926
7 0.186942386627197
8 0.189628330022097
9 0.167245028167963
10 0.169621634185314
11 0.152322239950299
12 0.154122185856104
13 0.143193371072412
14 0.14355327732861
15 0.138710526451468
16 0.13865713916719
17 0.13562601312995
18 0.13558938421309
19 0.133446518704295
20 0.133334505185485
21 0.131978513151407
22 0.131810494363308
23 0.130628834888339
24 0.130689896047115
25 0.129852763488889
26 0.129869377538562
27 0.129241408258677
28 0.129184298366308
29 0.128641336262226
30 0.128654422685504
31 0.128155825436115
32 0.128263763189316
33 0.128061847686768
34 0.128086376488209
35 0.127579492405057
36 0.12770385414362
37 0.127492285445333
38 0.127546788677573
39 0.127486929520965
40 0.127611554041505
41 0.12727059982717
42 0.127358559370041
43 0.12705399222672
44 0.127062376439571
45 0.126967048272491
46 0.127004427462816
47 0.126946412846446
48 0.126991738602519
49 0.126898736208677
50 0.126965687423944
51 0.168033064678311
52 0.169572784081101
53 0.165464062392712
54 0.167285034880042
55 0.164944484755397
56 0.166142778024077
57 0.164298317581415
58 0.164734856933355
59 0.145721501633525
60 0.145877421721816
61 0.142042352557182
62 0.141720451414585
63 0.138952074274421
64 0.138221651092172
65 0.136098674610257
66 0.135444506108761
67 0.133790386468172
68 0.133280432596803
69 0.132241926193237
70 0.13213249117136
71 0.131257769092917
72 0.130915319025517
73 0.130165640115738
74 0.129864085018635
75 0.129560668468475
76 0.12929760299623
77 0.128834445849061
78 0.128925909772515
79 0.128850549682975
80 0.128862247243524
81 0.128448681011796
82 0.128404958322644
83 0.128485916256905
84 0.128546485826373
85 0.128419828414917
86 0.128413282781839
87 0.128396026864648
88 0.12861519753933
89 0.128490808978677
90 0.128575296103954
91 0.128425786793232
92 0.128387218639255
93 0.128220211938024
94 0.128203415721655
95 0.128143819719553
96 0.128037516772747
97 0.127863150164485
98 0.127895519807935
99 0.1278422845155
100 0.127910759821534
};
\addlegendentry{$\cecw$}
\addplot [line width=1pt, black, dashed]
table {%
2 0.432405945360661
3 0.431935333609581
4 0.431613813340664
5 0.211106497198343
6 0.215220319181681
7 0.186951753497124
8 0.189657213538885
9 0.167238251119852
10 0.169611628502607
11 0.15232959382236
12 0.154094683378935
13 0.143139592185617
14 0.143508154526353
15 0.138698717430234
16 0.138659113571048
17 0.135604070425034
18 0.135587404742837
19 0.133430672958493
20 0.133318739756942
21 0.131996283084154
22 0.131812476366758
23 0.130621331557632
24 0.130683856904507
25 0.129832512065768
26 0.129873385205865
27 0.129253019690514
28 0.12919636875391
29 0.128659525960684
30 0.12868901334703
31 0.128176240772009
32 0.128280184715986
33 0.128090429753065
34 0.16987543143332
35 0.16945454351604
36 0.167088869959116
37 0.166739872321486
38 0.166423451751471
39 0.165542710050941
40 0.165528651177883
41 0.164648578464985
42 0.146616199836135
43 0.14531618617475
44 0.14216485440731
45 0.141435634568334
46 0.139218797460198
47 0.137705128192902
48 0.135870956853032
49 0.135056128874421
50 0.133824721574783
51 0.133254015147686
52 0.132478985562921
53 0.131903936266899
54 0.131183047443628
55 0.1308341114223
56 0.130435506552458
57 0.130254967361689
58 0.13001382984221
59 0.129751160144806
60 0.129504031315446
61 0.129513005241752
62 0.129460485130548
63 0.129402815699577
64 0.129402316436172
65 0.129311151430011
66 0.129022178053856
67 0.18441693149507
68 0.186100345700979
69 0.180464167222381
70 0.182667581662536
71 0.176861809119582
72 0.17718715198338
73 0.176704060137272
74 0.17795352101326
75 0.157026564031839
76 0.157599873766303
77 0.153633786663413
78 0.15266848385334
79 0.14982981748879
80 0.14949288032949
81 0.148548218458891
82 0.14878999568522
83 0.146786546781659
84 0.147215016111732
85 0.146804156079888
86 0.147680530175567
87 0.146279211416841
88 0.144811911061406
89 0.14393947057426
90 0.144674732536077
91 0.144802676960826
92 0.145387483909726
93 0.144095004796982
94 0.145016466900706
95 0.147259957641363
96 0.144824330732226
97 0.144253636077046
98 0.144154978469014
99 0.14327983379364
100 0.142914838343859
};
\addlegendentry{$\cecwdlr$}
\end{axis}

\end{tikzpicture}}
    \scalebox{.65}{
\begin{tikzpicture}

\definecolor{darkgray176}{RGB}{176,176,176}
\definecolor{green}{RGB}{0,128,0}
\definecolor{lightgray204}{RGB}{204,204,204}

\begin{axis}[
legend cell align={left},
legend style={fill opacity=0.8, draw opacity=1, text opacity=1, draw=lightgray204},
tick align=outside,
tick pos=left,
title={Carmon2019Unlabeled},
x grid style={darkgray176},
xlabel={iteration \(\displaystyle k\)},
xmajorgrids,
xmin=-2.9, xmax=104.9,
xtick style={color=black},
y grid style={darkgray176},
ylabel={\(\displaystyle ||\mathbf{x}^{(k+1)} - \mathbf{x}^{(k)}||_2\)},
ymajorgrids,
ymin=0.129969293322414, ymax=0.446708332080394,
ytick style={color=black}
]
\addplot [ultra thick, red]
table {%
2 0.432181698977947
3 0.431797012686729
4 0.43146017074585
5 0.228087570667267
6 0.233133678287268
7 0.203737695366144
8 0.207296037524939
9 0.185412873476744
10 0.188259896785021
11 0.172048420980573
12 0.174268977120519
13 0.163409437760711
14 0.164088761508465
15 0.158803266435862
16 0.159119322374463
17 0.156030699014664
18 0.155830314084888
19 0.153154429942369
20 0.152778103873134
21 0.150997102558613
22 0.151101902723312
23 0.150006836205721
24 0.149725984558463
25 0.148878066390753
26 0.149049386382103
27 0.148267648816109
28 0.148496676981449
29 0.147823035269976
30 0.147848484218121
31 0.147433515191078
32 0.147611290514469
33 0.147204326540232
34 0.147275486811996
35 0.147073062062263
36 0.146974182352424
37 0.146633715704083
38 0.146661293581128
39 0.146615369915962
40 0.146773799806833
41 0.146466695666313
42 0.146436103284359
43 0.145974014997482
44 0.146082903891802
45 0.145944282189012
46 0.146011448353529
47 0.146029826328158
48 0.145953708812594
49 0.145732834488153
50 0.145699463486671
51 0.145588280111551
52 0.145711957588792
53 0.145453447178006
54 0.145463169589639
55 0.14531498350203
56 0.145254561826587
57 0.145322764739394
58 0.14527348048985
59 0.14504748262465
60 0.145015513449907
61 0.145068596377969
62 0.145204498618841
63 0.145037946552038
64 0.14508163228631
65 0.145113704577088
66 0.145017259716988
67 0.144874258339405
68 0.144817437827587
69 0.144593767598271
70 0.14450132958591
71 0.144789934754372
72 0.145037962123752
73 0.14481742143631
74 0.144712406024337
75 0.144753767699003
76 0.144856055006385
77 0.144769869819283
78 0.144704304561019
79 0.144606141522527
80 0.144759308695793
81 0.144611869901419
82 0.144568603187799
83 0.144585581943393
84 0.14474905565381
85 0.144904615879059
86 0.144802778288722
87 0.144608076736331
88 0.14467877022922
89 0.144675957858562
90 0.144463031440973
91 0.144366522356868
92 0.144478017985821
93 0.144430756941438
94 0.1444139187783
95 0.144547250196338
96 0.14473512917757
97 0.144563078805804
98 0.144383153766394
99 0.144538277238607
100 0.144610862880945
};
\addlegendentry{$\ce$}
\addplot [ultra thick, green]
table {%
2 0.43231110304594
3 0.431823972463608
4 0.43149000287056
5 0.230960897654295
6 0.235878599435091
7 0.206213624179363
8 0.209539694935083
9 0.187648573815823
10 0.190479564219713
11 0.173639044389129
12 0.175772375389934
13 0.165179800018668
14 0.16588453181088
15 0.160183927044272
16 0.160615922138095
17 0.157204673886299
18 0.157125075012445
19 0.154918092340231
20 0.155009939447045
21 0.153128278404474
22 0.153283913061023
23 0.1521240696311
24 0.152090258300304
25 0.151008739992976
26 0.150889940112829
27 0.150429428592324
28 0.150467332005501
29 0.14995845168829
30 0.14988120034337
31 0.149335375279188
32 0.149386458098888
33 0.148693267926574
34 0.148801017776132
35 0.148632801771164
36 0.148581096604466
37 0.148204670101404
38 0.148452409207821
39 0.148126498684287
40 0.148257276564837
41 0.148242843747139
42 0.148198689743876
43 0.147804063931108
44 0.147973783537745
45 0.148048432469368
46 0.148008391484618
47 0.147770320102572
48 0.14770584538579
49 0.147725705355406
50 0.147669970095158
51 0.147573350593448
52 0.147647417709231
53 0.147486783117056
54 0.147533660680056
55 0.147517626807094
56 0.147627555578947
57 0.147628278508782
58 0.147723295018077
59 0.147466430962086
60 0.147308732196689
61 0.147294023036957
62 0.147441123202443
63 0.147375559434295
64 0.147265417650342
65 0.147159914970398
66 0.147298823669553
67 0.147069515287876
68 0.147040487229824
69 0.146971607431769
70 0.147151582539082
71 0.147061575725675
72 0.147091793715954
73 0.146893993094563
74 0.146894402205944
75 0.146823585629463
76 0.146866567805409
77 0.146960331574082
78 0.146786759421229
79 0.1467457690835
80 0.146858300715685
81 0.146806895062327
82 0.146722112596035
83 0.146623384729028
84 0.146654549986124
85 0.146578271090984
86 0.146647554188967
87 0.146741890981793
88 0.146731108948588
89 0.146630745306611
90 0.146685076653957
91 0.146713291332126
92 0.146610728800297
93 0.146331952363253
94 0.146377557888627
95 0.14647777326405
96 0.146306031942368
97 0.146269543170929
98 0.146528405770659
99 0.146525613218546
100 0.146454912126064
};
\addlegendentry{$\cw$}
\addplot [line width=1pt, blue, dashed]
table {%
2 0.432181698977947
3 0.431797012686729
4 0.43146017074585
5 0.228087570667267
6 0.233133678287268
7 0.203737695366144
8 0.207296037524939
9 0.185412873476744
10 0.188259896785021
11 0.172048420980573
12 0.174268977120519
13 0.163409437760711
14 0.164088761508465
15 0.158791893571615
16 0.159112831428647
17 0.156019118279219
18 0.155792439654469
19 0.153128148168325
20 0.152761620208621
21 0.151000437438488
22 0.151098617613316
23 0.149974074661732
24 0.149703997299075
25 0.148851285874844
26 0.149022243618965
27 0.14829660743475
28 0.148528650999069
29 0.147830478549004
30 0.147875076681376
31 0.147430713176727
32 0.147645948529243
33 0.147311005294323
34 0.147330006584525
35 0.147092610895634
36 0.147004436776042
37 0.146706339344382
38 0.146700798645616
39 0.146626534312963
40 0.146828379631042
41 0.146498265117407
42 0.146481822133064
43 0.145982689857483
44 0.146117533296347
45 0.146012278571725
46 0.14605778157711
47 0.146055160984397
48 0.145958474352956
49 0.145779986381531
50 0.14574608951807
51 0.186711798980832
52 0.189165019243956
53 0.184536237716675
54 0.186607843637466
55 0.184010224491358
56 0.185362215787172
57 0.183131372332573
58 0.183397543802857
59 0.16489461235702
60 0.164265372008085
61 0.160346578136086
62 0.159504401534796
63 0.156671563833952
64 0.156311555504799
65 0.154411834552884
66 0.153787945881486
67 0.151980189010501
68 0.151630560532212
69 0.150746585428715
70 0.150583337768912
71 0.149760768264532
72 0.149559017196298
73 0.14906201466918
74 0.148963588625193
75 0.148542632982135
76 0.148513740301132
77 0.148391985148191
78 0.148355023562908
79 0.147902498915792
80 0.147883300557733
81 0.147586233243346
82 0.147314569801092
83 0.147554603517056
84 0.147550922930241
85 0.147398177683353
86 0.147477301135659
87 0.147212913483381
88 0.147086919695139
89 0.146980815082788
90 0.1471136033535
91 0.14692185536027
92 0.146902864798903
93 0.146909447684884
94 0.146941828504205
95 0.146766221225262
96 0.146936306059361
97 0.146825136616826
98 0.146655711531639
99 0.146512741222978
100 0.146492914482951
};
\addlegendentry{$\cecw$}
\addplot [line width=1pt, black, dashed]
table {%
2 0.432181698977947
3 0.431797012686729
4 0.43146017074585
5 0.228087570667267
6 0.233133678287268
7 0.203737695366144
8 0.207296037524939
9 0.185412873476744
10 0.188259896785021
11 0.172048420980573
12 0.174268977120519
13 0.163409437760711
14 0.164088761508465
15 0.158793538212776
16 0.159114480987191
17 0.156024105399847
18 0.155799090489745
19 0.153128148168325
20 0.152761620208621
21 0.150998744368553
22 0.151098617613316
23 0.149972386211157
24 0.149691861495376
25 0.148823730349541
26 0.148993691056967
27 0.148261552155018
28 0.148509494364262
29 0.147807440161705
30 0.147850886285305
31 0.147406560629606
32 0.147578975111246
33 0.147225880622864
34 0.188932406976819
35 0.189637801647186
36 0.187160740792751
37 0.186674144938588
38 0.186126858144999
39 0.184902775138617
40 0.184517831429839
41 0.183197377249599
42 0.166047750115395
43 0.164547027871013
44 0.16111790984869
45 0.159732055515051
46 0.157444479167461
47 0.1562406937778
48 0.154909683987498
49 0.153751660734415
50 0.152179730981588
51 0.151825548410416
52 0.151127348840237
53 0.150576251745224
54 0.149934532865882
55 0.149378428906202
56 0.149115351587534
57 0.148827130123973
58 0.148422418683767
59 0.148331506028771
60 0.148261942565441
61 0.148119162544608
62 0.147920324206352
63 0.14777107000351
64 0.147800793126225
65 0.14770417816937
66 0.147460027337074
67 0.205098823159933
68 0.207225851416588
69 0.20248758777976
70 0.199256649166346
71 0.195619872957468
72 0.195540876835585
73 0.191685543805361
74 0.19198232807219
75 0.173432763516903
76 0.173782033622265
77 0.166283021792769
78 0.164489869251847
79 0.162864313423634
80 0.160053585618734
81 0.154479824900627
82 0.155466361939907
83 0.154003888592124
84 0.152465101480484
85 0.152196785435081
86 0.151612256243825
87 0.151259427070618
88 0.150405390411615
89 0.14969806663692
90 0.150583301559091
91 0.150345123261213
92 0.150459975525737
93 0.149788013473153
94 0.149666896611452
95 0.149895083829761
96 0.149805531650782
97 0.149559456929564
98 0.149366513043642
99 0.147423188760877
100 0.145115315020084
};
\addlegendentry{$\cecwdlr$}
\end{axis}

\end{tikzpicture}}
    \caption{Plotting the $\lt$-norm between successive PGD steps, for various surrogate losses. Each panel represents this quantity over iterations, for a different classifier (ModelID is on top of each panel). }
    \label{fig:qualitative}
\end{figure*}

\subsubsection{Multi-Stage PGD versus AutoAttack Components} Next, we compare our best method (on average, that is $\cecwdlr$) with every white-box component from AutoAttack \cite{croce2020reliable}, i.e., $\apgdce$, $\apgddlr$ and FAB attack \cite{Croce2020MinimallyDA}. In the original AutoAttack evaluation, the last two components are run for $T=100$ iterations and $R=9$ restarts, using the targeted version of each attack. However, we adapt these attacks to our computational budget, evaluating the performance of their untargeted versions for $T=100$. In our experiments, we execute the official code\footnote{https://github.com/fra31/auto-attack} of AutoAttack for every single model.  We clarify that the official code does not provide a way to turn off random initialization when evaluating the AA components, but the fluctuations are expected to be small enough. 

As it is clearly illustrated in \autoref{table:autoattack}, our proposed method, $\cecwdlr$ consistently outperforms the white-box components of AutoAttack. It becomes evident that the advantage of using the loss switching strategy is significant, since in this setting we run our attack for fixed step size equal to $\epsilon/4$ and the simplest optimizer possible (sign operation with no momentum). $\apgdce$ and $\apgddlr$ are both based in the evidently better APGD optimizer and step size is decayed according to some schedule, yet they lag behind $\cecwdlr$ by a large margin. Particularly, $\cecwdlr$ achieves (on average) 0.418\% lower robust accuracy than the strongest component.

\subsubsection{Multi-Stage PGD versus the strongest baselines} We extend the assessment of our method's effectiveness by comparing it with the strongest $\linf$-bounded attacks on CIFAR-10, for $T=100$ and no restarts. We consider the two best baselines found in literature (in our computational budget): GAMA-PGD \cite{Sriramanan2020GuidedAA} and MD attack \cite{Ma2020ImbalancedGA}. Both of these methods suggest improving PGD through modifications on the surrogate loss and step size schedule. In Subsection \ref{subsection:similarity}, we delve into the exact similarities between the examined methods and our work. 

We execute these attacks through the official codebases\footnote{https://github.com/val-iisc/GAMA-GAT} \footnote{https://github.com/Jack-lx-jiang/MD\_attacks}. When comparing $\cecwdlr$ with each baseline, we adapt the learning rate schedule according to each work (See \hyperref[appendix]{Appendix} for details). The results of these comparisons are summarized in \autoref{table:baselines}. In the parentheses of $\cecwdlr$ columns, we display which learning rate schedule is used. These results indicate the effectiveness of our attack, achieving state-of-the-art performance (in the $T=100,R=1$ budget), for the majority of evaluated models.

Specifically, $\cecwdlr$ outperforms GAMA-PGD \cite{Sriramanan2020GuidedAA} in 11 out of 15 $\linf$-bounded robust models, whereas in 2 models they achieve the exact same ASR. In the 2 networks that $\cecwdlr$ returns higher robust accuracy, the differences are quite small, i.e., $0.02\%$ and $0.04\%$. An extreme case is the model of \cite{zhang2020geometry}, since GAMA-PGD lags behind our method for $1.10\%$. These observations indicate that, in general, $\cecwdlr$ suffers less from robustness overestimation.  

In the case of MD attack \cite{Ma2020ImbalancedGA}, our method achieves lower robust accuracy in 13 out of 15 tested models, with an average improvement of $0.15\%$. In two models \cite{hendrycks2019using,sehwag2020hydra}, however, the estimated robust accuracy is $0.05\%$ and $0.14\%$ higher than that of MD attack. Overall, this comparison, similarly to the previous one, highlights that $\cecwdlr$ provides the most reliable $\linf$-bounded robustness evaluations. 

Since the differences of our best method with these baselines are marginal for some cases, it is crucial to answer whether they arise just because our method is benefited from the specific PGD starting point (which in our case is the clean image). To address this, we repeat the above comparisons for the same amount of iterations but with $R=5$ restarts. In this case, the starting points are initialized with random noise $\delta = \epsilon \cdot \text{sgn}(\vect{u})$, where $\vect{u} \sim \mathcal{U}(-1,1)$. The results are illustrated in \autoref{table:baselines_v2}. Overall, it is evident that the increased number of restarts helps each attack to achieve slightly lower robust accuracy, but comparatively, our attack still performs more reliable robustness evaluations for the vast majority of cases. 

\begin{table*}[]
\centering
\begin{tabular}{r  >{\bfseries}c c >{\bfseries}c c >{\bfseries}c c}
    \toprule
    Model ID & \normalfont{$\cecw$} & $\cwce$ & \normalfont{$\cedlr$} & $\dlrce$ & \normalfont{$\cwdlr$} & $\dlrcw$  \\
    \cmidrule(lr){1-1}  \cmidrule(lr){2-3} \cmidrule(lr){4-5} \cmidrule(lr){6-7} 
    
    Engstrom2019Robustness~\cite{robustness}             & 50.29 & 50.95 & 50.22 & 51.13 & 52.63 & 52.97 \\
    Carmon2019Unlabeled~\cite{carmon2019unlabeled}       & 60.00 & 60.27 & 60.00 & 60.39 & 60.88 & 60.94\\
    Hendrycks2019Using~\cite{hendrycks2019using}         & 55.41 & 55.62 & 55.37 & 55.72 & 56.55 & 56.84\\
    Zhang2019You~\cite{zhang2019you}                     & 45.33 & 45.85 & 45.32 & 45.86 & 47.42 & 47.58\\
    Zhang2019Theoretically~\cite{zhang2019theoretically} & 53.45 & 53.76 & 53.43 & 53.86 & 54.23 & 54.31\\
    Wu2020Adversarial~\cite{wu2020adversarial}           & 56.47 & 56.68 & 56.44 & 56.72 & 56.94 & 56.98\\
    Sehwag2021Proxy\_R18~\cite{sehwag2021robust}         & 56.06 & 56.42 & 56.05 & 56.52 & 57.21 & 57.49 \\
    Andriushchenko2020Understanding~\cite{andriushchenko2020understanding} & 44.56 & 44.94 & 44.53 & 45.03 & 46.62 & 46.77 \\
    Dai2021Parameterizing~\cite{dai2021parameterizing}   & 61.80 & 62.18 & 61.76 & 62.33 & 63.23 & 63.42 \\
    Gowal2021Improving\_28\_10\_ddpm\_100m~\cite{gowal2021improving} & 63.86 & 64.80 & 63.85 & 64.30 & 65.20 & 65.31 \\
    Huang2021Exploring\_ema\cite{huang2021exploring}         & 63.09 & 63.52 & 63.03 & 63.64 & 64.12 & 64.34 \\
    Zhang2020Geometry~\cite{zhang2020geometry}               & 59.78 & 60.16 & 59.69 & 60.31 & 60.37 & 60.51 \\
    Rade2021Helper\_R18\_extra~\cite{rade2021helperbased}    & 57.77 & 58.17 & 57.74 & 58.18 & 58.51 & 58.54 \\
    Addepalli2021Towards\_RN18~\cite{addepalli2021towards}   & 51.45 & 51.79 & 51.43 & 51.84 & 51.86 & 51.93 \\
    Sehwag2020Hydra~\cite{sehwag2020hydra}                   & 57.66 & 57.92 & 57.61 & 57.93 & 58.40 & 58.47 \\
    \bottomrule
    \end{tabular}
\vspace{0.2cm}
\caption{Ablation Study. Exploring the importance of the surrogates' order. The experiments are executed for $T=100$ with no restarts. Each entry reports the robust accuracy of each classifier for the given method. (\dag): Attacked with $\epsilon=0.031$.}
\label{table:ablation1}
\end{table*}
\begin{table*}[t]
\centering
\begin{tabular}{r c c c c c >{\bfseries}c}
    \toprule
             &       &       &                     &    \multicolumn{2}{c}{Convex} &                      \\
    Model ID & $\ce^{100}$ & $\cw^{100}$ & $\ce^{50}\lor\cw^{50}$ & $\gamma=0.25$ & $\gamma=0.75$ & \normalfont{$\cecw$}  \\ 
    \cmidrule(lr){1-1}  \cmidrule(lr){2-2} \cmidrule(lr){3-3} \cmidrule(lr){4-4} \cmidrule(lr){5-6} \cmidrule(lr){7-7}
    Engstrom2019Robustness \cite{robustness}                               & 52.24 & 52.59 & 50.75 & 51.59 & 52.30 & 50.29 \\
    Carmon2019Unlabeled \cite{carmon2019unlabeled}                         & 62.09 & 60.86 & 60.18 & 60.97 & 60.85 & 60.00 \\
    Hendrycks2019Using \cite{hendrycks2019using}                           & 57.38 & 56.61 & 55.52 & 56.10 & 56.36 & 55.41 \\
    Zhang2019You \cite{zhang2019you}                                       & 46.28 & 47.44 & 45.56 & 46.37 & 47.19 & 45.33 \\
    Zhang2019Theoretically \cite{zhang2019theoretically} \dag              & 55.47 & 54.21 & 53.68 & 54.34 & 54.18 & 53.45 \\
    Wu2020Adversarial \cite{wu2020adversarial}                             & 59.05 & 56.93 & 56.66 & 57.46 & 57.00 & 56.47 \\
    Sehwag2021Proxy\_R18\cite{sehwag2021robust}                            & 58.68 & 57.22 & 56.37 & 57.22 & 57.10 & 56.06 \\
    Andriushchenko2020Understanding \cite{andriushchenko2020understanding} & 47.14 & 46.62 & 44.81 & 45.78 & 46.17 & 44.56 \\
    Dai2021Parameterizing \cite{dai2021parameterizing}                     & 63.98 & 63.23 & 62.17 & 62.89 & 63.15 & 61.80 \\
    Gowal2021Improving\_28\_10\_ddpm\_100m\cite{gowal2021improving}        & 65.79 & 65.20 & 64.20 & 64.72 & 65.00 & 63.86 \\
    Huang2021Exploring\_ema \cite{huang2021exploring}                      & 64.95 & 64.15 & 63.35 & 64.01 & 64.09 & 63.09 \\
    Zhang2020Geometry \cite{zhang2020geometry}                             & 66.67 & 60.40 & 60.14 & 63.88 & 60.86 & 59.78 \\
    Rade2021Helper\_R18\_extra \cite{rade2021helperbased}                  & 61.48 & 58.51 & 58.14 & 59.14 & 58.49 & 57.77 \\
    Addepalli2021Towards\_RN18 \cite{addepalli2021towards}                 & 56.00 & 51.88 & 51.78 & 53.23 & 51.96 & 51.45 \\
    Sehwag2020Hydra \cite{sehwag2020hydra}                                 & 59.86 & 58.41 & 57.85 & 58.59 & 58.41 & 57.66 \\
    \bottomrule
    \end{tabular}
\vspace{0.2cm}
\caption{Ablation Study. In the convex columns, $\gamma$ ($1-\gamma$) corresponds to CE (CW). The experiments are executed for $T=100$ with no restarts. Each entry reports the robust accuracy of each classifier for the given method. (\dag): Attacked with $\epsilon=0.031$.}
\label{table:ablation2}
\end{table*}

\subsection{Qualitative Analysis}

Here, we conduct a qualitative analysis to better grasp the impact of changing surrogate losses during optimization. Our experiments are inspired by the work of Yamamura et al. \cite{yamamura2022diversified}, where they visualize the $\lt$-distance between successive PGD steps: $\|\vx^{(k+1)}-\vx^{(k)}\|_2$ in order to empirically show that their proposed optimizer explores the input space more extensively. In a similar vein, we replicate their method for $\ce,\cw,\cecw,\cecwdlr$ in \autoref{fig:qualitative}, inspecting three different classifiers. To generate smoother curves, the y-axis quantity is averaged on a batch of 100 examples.

Altogether, it appears that in the single loss variants, the search of PGD becomes quite localized and after some time the successive steps are within small distances. In the cases where multiple surrogates are used, the curve presents a sudden rise in the alternation timestep, indicating that the objective alternation helps the algorithm to diversify its search. 

\subsection{Ablation: Surrogate Loss Order in Multi-Stage PGD} 

A research question regarding the multi-stage variant of PGD is whether the objective ordering affects the results. Specifically, we are interested in understanding whether any change occurs if we optimize the objectives with reverse ordering. To address this question, we execute the two-stage PGD, with $T=100$ and no restarts, for every possible pair (order matters) of CE, CW and DLR. 

The results of \autoref{table:ablation1} demonstrate that the order plays an essential role. Particularly, it is clearly illustrated that it is better to start the optimization procedure with the CE loss, then finishing off with CW or DLR. 
However, we observe that regardless of the objective ordering, every multi-stage PGD variant which alternates between CE and one of CW, DLR ($\cecw$, $\cwce$, $\cedlr$, $\dlrce$ columns) performs better than single-loss PGD.

\subsection{Ablation: Additional Techniques of Combining Surrogates}

Another interesting research question is to explore whether there exist other ways of combining surrogates. To settle this, we compare the alternation method with two additional combining techniques.
First, one can combine different surrogates through a convex combination, i.e., setting the surrogate according to the following expression:
$$ \cL(\vx,y) = \gamma \cdot \cL_1(\vx,y) + (1-\gamma) \cdot \cL_2(\vx,y) $$
Another way is to combine different surrogates in an ensemble-like manner, i.e., split the entire iteration budget into $K$ equally sized intervals, execute PGD using the $k$-th surrogate $\cL_k$, starting from the clean point (not from where the previous stage ended), and then aggregate the output decisions. This method is inspired by the MultiTargeted surrogate, introduced by Gowal et al. \cite{Gowal2019AnAS}. For the CE and CW losses, we denote the latter combining strategy as $\ce\lor\cw$, because the output decisions of each surrogate are aggregated through binary OR, i.e., the input is deemed misclassified if at least one of $\ce$, $\cw$ generate a successful perturbation.  

We conduct an ablation study, using the CE and CW objectives, to explore the effectiveness of these methods. The results are illustrated in \autoref{table:ablation2}, where we also report the robust accuracy of $\ce,\cw,\cecw$ for direct comparison (we also include the iteration budget on the superscript to draw a distinction with the ensemble method). As expected, the robust accuracy of convex combination is susceptible to the choice of $\gamma$, with its performance depending on whether the best-performing objective has a larger weight. The ensemble method, on the other hand, consistently outperforms the single-loss PGD, and much like $\cecw$, is more ``robust" against issues arising from individual use of objectives. However, the loss alternation strategy, $\cecw$, performs better than the ensemble-like combination. We advocate that this occurs because $\cecw$ utilizes the progress made in previous stages to perform better initialization for the next stage. The ensemble-like method, however, discards the perturbation found by previous objectives, and starts optimization all over again. 
\section{Discussion} \label{sec:conclusion}

\subsection{Similarity with Previous Works}\label{subsection:similarity}

Next, we discuss previous works that also employ a loss alternating strategy. 
First, the most similar work is that of Ma et al. \cite{Ma2020ImbalancedGA}, where they employ an identical alternation step to evade the issue of imbalanced gradients. The first PGD stage optimize only one of the two logit terms, whereas in the final stage, the typical margin loss is optimized. Notice a striking difference: Our work involves the CE,CW and DLR losses, all containing more than one logit terms, hence potentially suffering from gradient imbalance that should translate to reduced ASR. Our method outperforms MD attack. Therefore, our study implies that the performance improvement of MD attack \cite{Ma2020ImbalancedGA} may be the outcome of switching surrogates, rather than deterring the magnitudes of logit terms' gradients from becoming highly disparate.

The second method is GAMA-PGD, introduced by Sriramanan et al. \cite{Sriramanan2020GuidedAA}. The authors propose to regularize the margin loss with a MSE term, weighted by a decaying coefficient. In their implementation, the initial rate of weights between the MSE and CW losses is 50:1, hence for the first few iterations the contribution of CW loss is negligible. The weight of MSE is linearly decayed to 0 for $T/4$ iterations, and after that point the surrogate is set to the standard margin loss. Essentially, their attack alternates the surrogate loss used by PGD as many times as the duration of the interval during which MSE decays, i.e., $T/4$ out of $T$ iterations. Their analysis conveys the intuition that the improvement originates solely from the regularizing effect that MSE exerts on the margin loss. Our work demonstrates that the benefits of GAMA-PGD may arise from the loss alternation, still further experimentation is required.

Another method loosely connected with ours is the Composite Adversarial Attack (CAA) \cite{mao2021composite}. Mao et al. propose to generate adversaries by searching for the best composition of individual base attacks. Our method can be seen as a more special study of CAA, since it composes PGD attacks for two (or three) different objectives. Our work indicates much more markedly the value of using multiple losses. The effectiveness of CAA appears more like the result of a brute-force-like search. 

Overall, our paper differs from the aforementioned works in that it manages to showcase the true efficacy of the alternation step, stripped down from other redundant components. The experiments provide direct evidence that using multiple objectives is sufficient to induce large performance gains. Additionally, our work is an extension of these methods since we evaluate the combination of all possible pairs of CE,CW and DLR losses, rather than using only CW with its individual terms \cite{Ma2020ImbalancedGA} or CW and MSE \cite{Sriramanan2020GuidedAA}. 

\subsection{Future Work}

There are several questions arising from the proposed work than require further investigation and could be of value to the community. Notably, it is critical to address whether there is a trade-off between the number of surrogates used and PGD performance, for a fixed number of iterations. We assumed that adding more stages for fixed budget may hinder performance due to the decreased duration allotted to each stage. However, our intuition is that adding more objectives shouldn't drop the Attack Success Rate (ASR), given that PGD spends a sufficient time in each stage. This can be easily verified by increasing the computational budget and observing wether the larger amount of surrogates leads to higher ASR. 

Another interesting observation to explore is how the alternation step depends on the choice of objectives and their respective formulations. Particularly, we observed that $\cwdlr$ performs at a par (or even worse) than the respective single-loss variants, $\cw$ and $\dlr$, which was credited to the similarity of CW and DLR. This indicates that the loss alternation technique is an improvement only if the expressions generate landscapes which are diverse enough. In this vein, it would be valuable to encompass other expressions which deviate from the objective functions of our study, i.e., CE, CW and DLR.  

Since we experimentally demonstrate that our PGD variant is the strongest adversarial attack in the computational budget of 100 iterations, another direct extension is to integrate our attack into powerful ensembles. Specifically, in the case of AutoAttack \cite{croce2020reliable}, $\cecwdlr$ is outperforming every white-box component (\autoref{table:autoattack}), hence we assume that replacing e.g. $\apgddlr$ with $\cecwdlr$ would produce more reliable robustness evaluations. 

Apart from that, it is worthwhile to investigate whether the idea of increasing the number of surrogates helps other algorithms to perform better. Notice that our work is entirely framed within the PGD algorithm, but other popular attacks remain unexplored. Subsequent works could address whether our findings extrapolate to other attacks, and even in other settings e.g. black-box attacks.  

\section{Conclusion}

In this work, we propose a method of alternating objectives for improving the strength of PGD-based attacks. The proposed method performs better than single loss variants, across 25 adversarial defenses, spanning 3 different datasets. In the CIFAR-10 case, it performs better than strong baselines which are used for evaluating the $\lp-$bounded robustness of neural networks: AutoPGD \cite{croce2020reliable}, FAB \cite{Croce2020MinimallyDA}, GAMA-PGD \cite{Sriramanan2020GuidedAA} and MD Attack \cite{Ma2020ImbalancedGA}.
Our experiments show that alternating objectives is a very effective way of combining different objectives compared, e.g., to convex combination and ensemble-like methods. It is also experimentally shown that the proposed method offers significant robustness towards overcoming loss-specific weaknesses. 
Furthermore, our qualitative analysis offers intuition on the reasons behind our method's strength that may be related to the algorithm's search space diversification induced by the alternation step. Finally, we offer a new perspective on how the success of other state-of-the-art attacks, i.e., GAMA-PGD and MD Attack, can be ascribed to loss alternation.

\bibliographystyle{IEEEtran}
\bibliography{main.bbl}

\clearpage 
\section*{Appendix}\label{appendix}

\subsection{Implementation Details}

For our experiments, we implement code on the PyTorch framework. The PGD implementation is based on the TRADES \cite{zhang2019theoretically} repository\footnote{https://github.com/yaodongyu/TRADES}. All attacks are executed with a $\linf$-norm bound of $\epsilon=8/255$ and for $T=100$ iterations, with no restarts. Our code returns the best intermediate PGD point instead of the last. The robust models of our study are obtained from the ModelZoo of RobustBench \cite{croce2021robustbench}. Our experiments are run in a NVIDIA GeForce GTX 1080 Ti GPU with 12GB VRAM. 

\subsection{Step Size Schedules}

Here, we discuss the step size schedules used when comparing our method with the GAMA-PGD \cite{Sriramanan2020GuidedAA} and MD Attack \cite{Ma2020ImbalancedGA} baselines. In GAMA-PGD, the step size schedule incurs tenfold drops at $T=60$ and $T=85$, starting from $\eta^{(0)} = 2\epsilon$. 

In \cite{Ma2020ImbalancedGA}, step size is regulated according to a cosine-annealing scheme. In particular, the step size in $t-th$ iteration equals:

\begin{equation*}
    \eta^{(t)} = \begin{cases}
    \epsilon \cdot (1 + \cos (\frac{t-1}{T'})\pi)  &, t < T'\\ 
    \epsilon \cdot (1 + \cos (\frac{t-T'}{T-T'}\pi)) &, T' \leq t < T  
    \end{cases}
\end{equation*}
where $T=100$, $T'=T/2$. Therefore, step size is decayed from $2\epsilon$ to 0 in each stage. We extend this scheme to our three-stage variant as follows:
\begin{equation*}
    \eta^{(t)} = \begin{cases}
    \epsilon \cdot (1 + \cos (\frac{t-1}{T/3})\pi)  &, t < T/3\\ 
    \epsilon \cdot (1 + \cos (\frac{t-T/3}{T/3}\pi)) &, T/3 \leq t < 2T/3 \\
    \epsilon \cdot (1 + \cos (\frac{t-2T/3}{T/3}\pi)) &, 2T/3 \leq t < T \\
    \end{cases}
\end{equation*}



\end{document}